\documentclass[lettersize,journal]{IEEEtran} % CAMERA READY VERSION
\usepackage{amsmath,amsfonts}
\usepackage{algorithmic}
\usepackage{algorithm}
\usepackage{array}
\usepackage[caption=false,font=normalsize,labelfont=sf,textfont=sf]{subfig}
\usepackage{textcomp}
\usepackage{stfloats}
\usepackage{url}
\usepackage{verbatim}
\usepackage{graphicx}
\usepackage{cite}
\usepackage{makecell}
\usepackage{flushend}
\usepackage{svg}
\usepackage[printonlyused,withpage,nolist]{acronym}
\usepackage{lipsum} 
\definecolor{uhhstone}{RGB}{59,81,91}
\usepackage{url} 
\usepackage[colorlinks,
pdfpagelabels,
pdfstartview = FitH,
bookmarksopen = true,
bookmarksnumbered = true,
linkcolor = uhhstone,
plainpages = false,
hypertexnames = false,
citecolor = uhhstone]{hyperref}
\usepackage[capitalise,noabbrev]{cleveref}
\usepackage{orcidlink}
\Crefname{figure}{Fig.}{Figs.}% {<type>}{<singular>}{<plural>}
\Crefname{equation}{Eq.}{Eqs.}% {<type>}{<singular>}{<plural>}
\usepackage{graphicx}
\def\BibTeX{{\rm B\kern-.05em{\sc i\kern-.025em b}\kern-.08em
    T\kern-.1667em\lower.7ex\hbox{E}\kern-.125emX}}
\usepackage[font=small,labelfont=bf,tableposition=top]{caption}

\usepackage{blindtext}

\usepackage{booktabs}
\usepackage{multirow}

\begin{document}
\begin{acronym}
    \acro{ais}[AIS]{Automatic Identification System}
    \acro{lidar}[LiDAR]{Light Detection and Ranging}
    \acro{radar}[RADAR]{Radio Detection and Ranging}
    \acro{bonk-pose}[BONK-Pose]{Boats on Nordelbe Kehrwieder}
    \acro{iou}[IoU]{intersection over union}
    \acro{map}[mAP]{mean average precision}
    \acro{ap}[AP]{average precision}    
    \acro{gps}[GPS]{Global Positioning System}
    \acro{yolo}[YOLO]{You Only Look Once}
    \acro{pnp}[PnP]{Perspective n-point}
    \acro{pca}[PCA]{Principle Component Analysis}
    \acro{dr}[DR]{Dead Reckoning}
    \acro{cog}[COG]{Course over Ground}
    \acro{add-s}[ADD-S]{Average Distance-Symmetric}
    \acro{detr}[DETR]{DEtection TRansformer}
\end{acronym}
%\linenumbers

\title{Fusing Monocular RGB Images with AIS Data to Create \\a 6D Pose Estimation Dataset for Marine Vessels}

\author{Fabian Holst*$^{,1}$ \orcidlink{0009-0007-1785-9371}, Emre Gülsoylu*$^{,1}$ \orcidlink{0000-0002-3834-3645}, Simone Frintrop$^{1}$ \orcidlink{0000-0002-9475-3593} %~\IEEEmembership{Staff,~IEEE,}
        % <-this % stops a space
% \thanks{$^{1}$University of Hamburg, Faculty of Mathematics, Informatics and Natural Sciences, Department of Informatics, Computer Vision Group, 22527, Hamburg, Germany. $\ast$ denotes equal contribution.}% <-this % stops a space
% \thanks{Manuscript received DD.MM.YYYY; revised DD.MM.YYYY.}
% }

\thanks{$^{1}$University of Hamburg, Faculty of Mathematics, Informatics and Natural Sciences, Department of Informatics, Computer Vision Group, 22527, Hamburg, Germany. .}% <-this % stops a space
\thanks{$\ast$ denotes equal contribution}
}

% \markboth{Journal of Oceanic Engineering,~Vol.~X, No.~X, MONTH~YYYY}%


\maketitle

\begin{abstract}
% This work addresses the challenge of 6D pose estimation, which involves determining an object's precise position and rotation from a single RGB image. This technology is crucial for applications in robotics and autonomous vehicles, enabling cost-effective object detection and localization using only a camera. We focus on the maritime domain, where 6D pose estimation can enhance the capabilities of autonomous vessels and improve traffic control. Our research details a setup for collecting images along a waterway while simultaneously acquiring AIS messages. We employ an object detector network to identify vessels in these images and develop methods to align real-world AIS data with image-based detections in a consistent coordinate system. This alignment is framed as a matching problem between predictions and AIS data, leading to a data fusion technique that integrates these elements into per-vessel 6D poses. We evaluate the effectiveness of various object detector networks, compare different transformation methods between image and world coordinates, and assess the performance of our AIS matching algorithm. Our efforts culminate in the creation of a novel dataset comprising 3753 annotated images, significantly contributing to the advancement of 6D pose estimation in maritime contexts.

The paper presents a novel technique for creating a 6D pose estimation dataset for marine vessels by fusing monocular RGB images with Automatic Identification System (AIS) data. The proposed technique addresses the limitations of relying purely on AIS for location information, caused by issues like equipment reliability, data manipulation, and transmission delays. By combining vessel detections from monocular RGB images, obtained using an object detection network (YOLOX-X), with AIS messages, the technique generates 3D bounding boxes that represent the vessels' 6D poses, i.e. spatial and rotational dimensions. The paper evaluates different object detection models to locate vessels in image space. We also compare two transformation methods (homography and Perspective-n-Point) for aligning AIS data with image coordinates. The results of our work demonstrate that the Perspective-n-Point (PnP) method achieves a significantly lower projection error compared to homography-based approaches used before, and the YOLOX-X model achieves a mean Average Precision (mAP) of 0.80 at an Intersection over Union (IoU) threshold of 0.5 for relevant vessel classes. We show indication that our approach allows the creation of a 6D pose estimation dataset without needing manual annotation. Additionally, we introduce the Boats on Nordelbe Kehrwieder (BONK-pose), a publicly available dataset comprising 3753 images with 3D bounding box annotations for pose estimation, created by our data fusion approach. This dataset can be used for training and evaluating 6D pose estimation networks. In addition we introduce a set of 1000 images with 2D bounding box annotations for ship detection from the same scene.
\end{abstract}

\begin{IEEEkeywords}
6D Pose Estimation, Automated Identification System, Ship Detection, Data Fusion.
\end{IEEEkeywords}

\section{Introduction}
\label{sec:intro}
The \ac{ais} aids in collision avoidance and enhances situational awareness for vessels and waterway management authorities \cite{USCG_AIS_FAQ} by providing information on vessels' position, course, and speed among other fields. Relying purely on \ac{ais} data for localisation has limitations, induced by equipment reliability issues \cite{emmens2021promises, tsou2010discovering}, manipulation \cite{mazzarella2015sar}, and delayed transmission that can lead to position drift. 

\begin{figure}[]
	\centering
	\includegraphics[width=\linewidth, trim=2cm 1cm 1cm 4.5cm, clip]{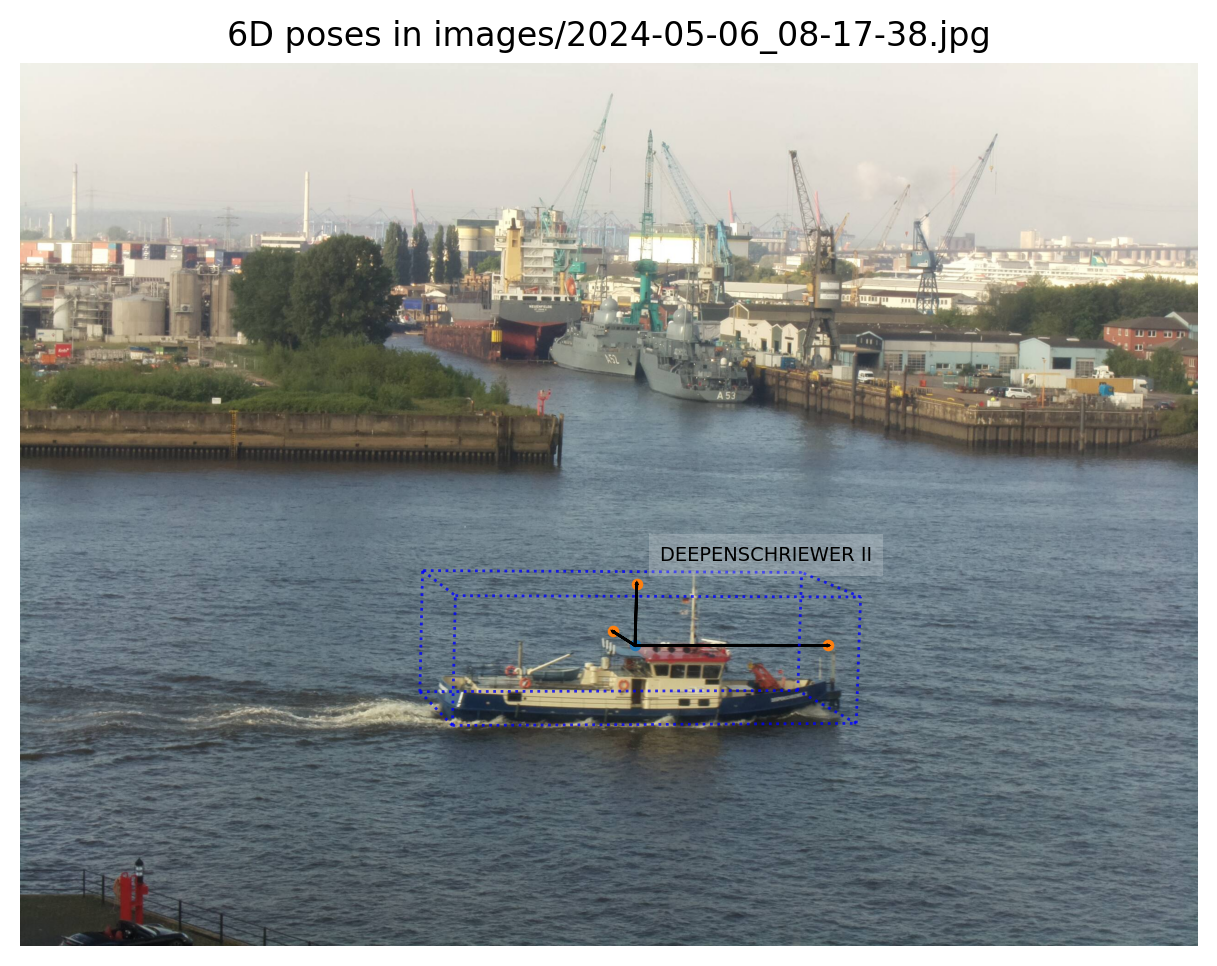}
	\caption{A sample visualisation of a 6D pose annotation, automatically created by fusion of \ac{ais} data and a predicted 2D bounding box. The 3D bounding box (blue dotted lines) encompasses the vessel and is aligned with the vessel's dimensions. The black axes with orange dots at the end represents the vessel's coordinate axes.}
	\label{fig:6DposeVisualized}
\end{figure}

To overcome the limitations of relying only on \ac{ais}, researchers have explored associating vessel instances (bounding boxes) with their respective \ac{ais} messages \cite{murray2021ais, zhang2025deep, lu2021fusion, solano2021detection, qu2022intelligent, carrillo2022ship, gulsoylu2024image}. By integrating visual information from monocular RGB images with \ac{ais} data, these approaches improve identification and tracking of vessels \cite{gao2025ship}, particularly in challenging environments where \ac{ais} signals may be incomplete, noisy, or subject to malicious modification. Despite these advancements, a critical challenge remains open: the accurate estimation of a vessel’s six degrees of freedom (6D) pose, that is, determining both its three spatial (position) and three rotational (orientation) dimensions. These information can be inferred through \ac{ais} data and monocular RGB image fusion, as demonstrated by this paper. Achieving accurate 6D pose estimation is challenging due to factors like occlusions, sensor noise, varying lighting conditions, and scene clutter, especially in dynamic environments \cite{marullo20236d}.

These challenges are not unique to the maritime field and the problem of 6D pose estimation has been extensively studied in other domains such as robotics \cite{cao20236impose, marullo20236d} and autonomous vehicles \cite{geiger2012we}. Similar to the challenges faced in autonomous car navigation, the accurate localisation and tracking of vessels in maritime environments is a critical task for ensuring safety and efficiency in waterway management \cite{reggiannini2024remote}. Leveraging 6D pose estimation, vessels can be represented as 3D bounding boxes, allowing for the estimation of their full spatial dimensions. This can complement the missing dimension information in \ac{ais}, which does not include the air draught information. Implementing 6D pose estimation would enrich the data for better localisation of vessels for waterway management and enable the detection of dimension-related disparities for surveillance.

While systems like \ac{lidar} or \ac{radar} offer point clouds on which the 6D pose of a vessel can be estimated \cite{hu2022estimation}, they are more expensive or have restricted ranges compared to cameras \cite{teixeira2022literature} and are more suitable for berthing assistance systems \cite{wang2024berthing}. Predicting the 6D pose of a vessel by using an RGB camera is important, as it would assist collision avoidance for autonomous ships and provides insights into vessel activities for traffic service and surveillance, while being much simpler and cheaper than \ac{lidar} or \ac{radar}.

\begin{figure*}
	\includegraphics[width=\linewidth]{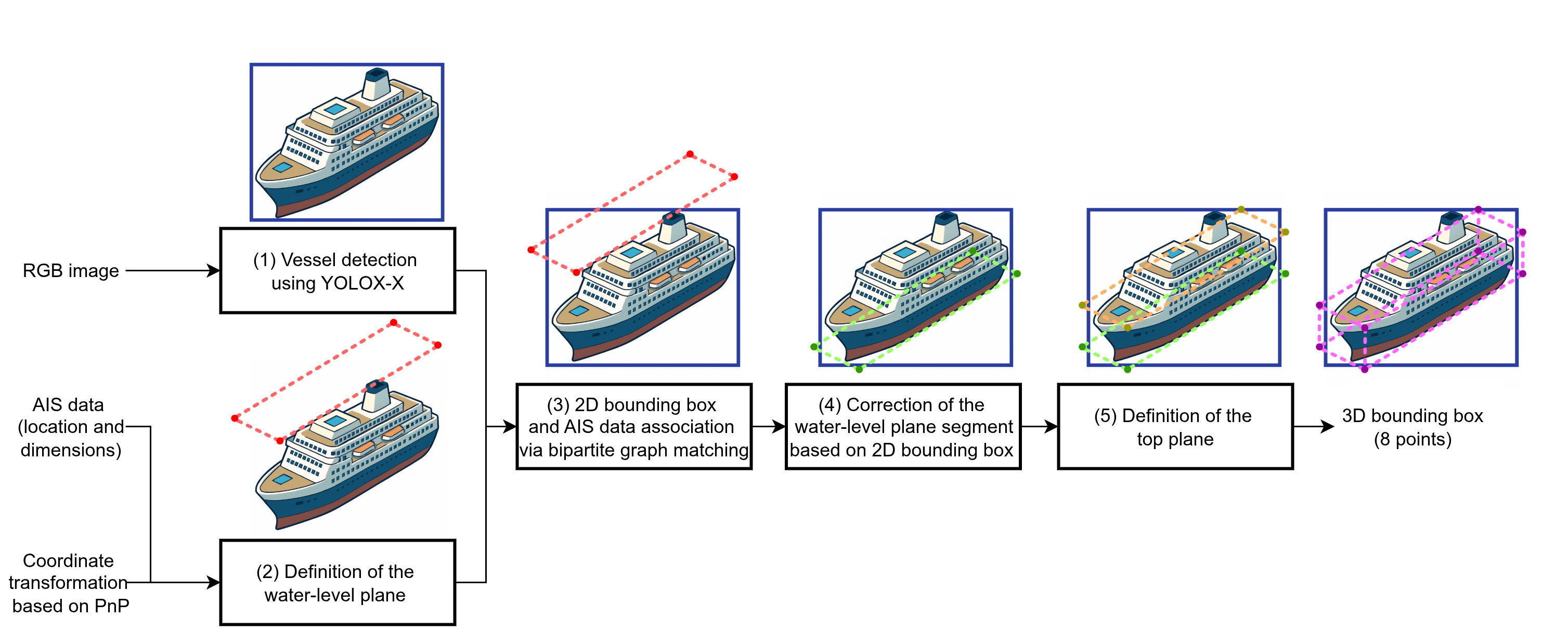}
	\caption{Flowchart of the proposed technique for creating 3D bounding boxes from \ac{ais} data and RGB image fusion. The inputs are the RGB image, \ac{ais} data, and the coordinate transformation matrix obtained via the \ac{pnp} algorithm: (1) Vessel detection using YOLOX-X and coordinate transformation of \ac{ais} data into image space based on the \ac{pnp} algorithm. (2) Definition of the water-level plane segment using the vessel's location and dimensions. (3) Association of 2D bounding boxes and \ac{ais} data through bipartite graph matching. (4) Correction of the water-level plane segment position based on the associated 2D bounding box. (5) Translation of the water-level plane segment to create the top edge of the 2D bounding box, resulting in a 3D bounding box for the detected vessel.}
	\label{fig:systemoverview}
\end{figure*}

%@Emre added this into sentence, i think before there was a sentence about maritime datasets having no full 6D pose esitmation dataset
So far there are no 6D pose estimation datasets specific to the maritime domain. In this work, we are addressing this gap by proposing a technique to create 3D bounding boxes using \ac{ais} data and monocular RGB images. Our technique involves detecting vessels in images using an object detection network, while simultaneously acquiring \ac{ais} messages. These detections are then associated with the \ac{ais} messages via bipartite graph matching. Finally, a 3D bounding box is created by fusing the two data sources as visualised in \cref{fig:6DposeVisualized}. The overview of the proposed technique is shown in \cref{fig:systemoverview}. We evaluate the effectiveness of various "off-the-shelf" object detection models and compare different transformation methods between image and world space. These methods are homography, as it is frequently used in the literature, and \ac{pnp} \cite{lepetit2009ep}, which has not been used for this task before. We also assess the performance of our \ac{ais} matching algorithm and 3D bounding box creation algorithm. 

Using the proposed technique, we created a novel and publicly available 6D pose estimation dataset, \ac{bonk-pose}, comprising 3753 annotated images. This dataset consists of RGB images taken by the River Elbe, relevant \ac{ais} messages, and annotations for training 6D pose estimation or ship detection networks. A sample image with annotations is shown in \cref{fig:6DposeVisualized}. 

Our goal is to allow machine learning based 6D pose estimation in the maritime domain, by both providing a dataset with ground truth and introducing a method to create further such datasets in a way that is not bottlenecked by manual annotation work.

We can summarise our contributions as follows:
\begin{itemize}
    \item A novel, automated technique for generating a 6D pose estimation dataset for marine vessels,
    \item A new and publicly available dataset\footnote{\href{https://fabianholst.github.io/BONK-pose/}{https://fabianholst.github.io/BONK-pose/}}, \acf{bonk-pose}, for ship detection and 6D pose estimation,
    \item Detailed evaluation of the proposed technique.
\end{itemize}

\section{Related work}
\label{sec:related_work}
There are already numerous studies focused on the fusion of RGB images with \ac{ais} data. This fusion requires not only  detecting vessels in images but also accurately associating each detected vessel with its transmitted \ac{ais} messages based on spatial and temporal information.

Lu et al. \cite{lu2021fusion} addressed this by using a modified YOLOv5 \cite{glenn_jocher_2022_yolov5} model to detect vessels in images from a shore-based camera overlooking the English Channel. By employing \ac{dr} to predict vessel positions at the image time and matching based on distance and bearing thresholds, they associated 72\% of message-vessel pairs. Limitations were due to inaccuracies in object detection and distance estimation, especially when vessel orientations did not match the assumptions made in the approach.

Huang et al. \cite{huang2021identity} combined \ac{ais} data with vessel tracking in inland waterways. They used DeepSORT \cite{Wojke2017simple} for tracking, with daytime objects detected by a custom SSD-based model and nighttime objects by detecting onboard lighting. \ac{ais} positions were interpolated using \ac{dr}. They applied homography to map image coordinates to real-world coordinates, enabling spatial distance calculations for matching. 

Carrillo-Pérez et al. \cite{carrillo2022ship} employed similar techniques but used pixel-wise segmentations and transformed image coordinates to world coordinates via homography. They also introduced a novel dataset called ShipSG, designed for the segmentation and georeferencing of ships in maritime monitoring scenes captured from static oblique views.

Extending \cite{carrillo2022ship} for panning cameras, Gülsoylu et al. \cite{gulsoylu2024image} aimed to enrich basic object detection outputs with detailed \ac{ais} information like vessel type and course. They fine-tuned YOLOv5 to detect vessels in images from public webcams. By estimating a homography between image and world spaces, they associated detected ships with \ac{ais} messages.
%\todo{There are some more papers which solve this problem with some more variance Qu2023, BallinesBarrera2023, Ding2024, should i add them?}

While current research offers a strong base for detecting and associating vessels, there is still a significant gap in the current research. None of the existing works have fused \ac{ais} data with RGB images to create a full 3D bounding box. This is mainly because \ac{ais} data does not include height information, requiring additional processing and new techniques to determine ship heights. Addressing this challenge could benefit from examining advancements in other fields. %Is a repitition of the next sentence
%, such as autonomous driving, where similar problems have been tackled with innovative solutions.

Similar to the maritime domain, detecting objects such as cars and people and estimating their 3D positions are crucial for decision making in autonomous driving \cite{ma20233d}. Additionally, when estimating the pose of objects such as humans or cars in traffic, the rotational axes are constrained in a manner similar to that of vessels in water \cite{kiefer2023stable}. These similarities create a potential for applying methods between domains. For example, Mousavian et al. \cite{mousavian20173d} address 6D pose estimation by first detecting 2D bounding boxes and estimating object dimensions based on category statistics. Combining these dimensions with rotation estimates allows them to project accurate 3D bounding boxes into image space, refining object positions within detections. 

In the maritime domain, Kiefer et al. \cite{kiefer2023stable} focus on estimating the yaw of vessels while assuming minimal roll and pitch. They enhance the HyperPosePDF method\cite{hofer2023hyperposepdf}, which uses a hypernetwork to generate probability distributions over possible rotations based on input images. By incorporating temporal information from multiple video frames, they improve accuracy by averaging out errors over time, resulting in more stable and precise yaw estimations. 

Autonomous driving has several domain specific datasets for 6D pose estimation. KITTI \cite{geiger2012we} is an early and frequently used dataset. It provides \ac{lidar} point clouds in addition to stereo images and a multitude of other sensor data. Another dataset is ApolloCar3D \cite{song2019apollocar3d}, which contains 5,277 images with over 60,000 instances of cars registered with their 3D CAD models. The 3D models are aligned with images by selecting keypoint matches between the models and visible parts of the vehicles. Annotations include the pose of each car in terms of translation and rotation relative to the camera.

Despite these advancements in autonomous driving, there remains a notable gap in the maritime domain. There are datasets available for tasks such as ship detection, including general purpose datasets like MS COCO \cite{lin2014microsoft} with their "boat" class or more tailored datasets to the maritime domain \cite{varga2022seadronessee}, as well as for instance segmentation \cite{carrillo2022ship}, and ship tracking \cite{qu2023improving}. However, there are few works focused on 6D pose estimation for ships \cite{ferreira20216d, kiefer2023stable}, and none that leverage \ac{ais} data. This gap will be filled by our work.

\section{Methodology}
We propose a technique to create 3D bounding boxes by fusing \ac{ais} data with 2D bounding boxes. The proposed technique consist of five stages, as shown in \cref{fig:systemoverview}: (1) The vessels in the image are detected by YOLOX-X \cite{ge2021yolox} and the location information gathered from \ac{ais} data is transformed from world coordinates into image coordinates. This transformation is achieved through a camera model whose parameters are determined by the \ac{pnp} algorithm. (2) The output of these operations, 2D bounding box predictions and available \ac{ais} data, are associated by bipartite graph matching. (3) Four points that encloses the vessel from bird's eye view are derived based on dimension information gathered from the \ac{ais} messages. These are obtained based on location, heading and vessel dimensions from \ac{ais} transformed into image space. The plane segment formed by these four corners is referred to as the water-level plane segment, which forms the bottom plane segment of the 3D bounding box. (4) As \ac{ais} is not completely accurate, as discussed in the \cref{sec:intro}, this plane segment may not perfectly fit the actual pose of the vessel. Thus the 2D bounding box predicted by YOLOX-X serves as a guide to correct the position of the water-level plane segment. (5) The top plane segment of the 3D bounding box is created by translating the water-level plane segment upwards, using the top edge of the 2D bounding box as a restraint. The result of this technique is a 3D bounding box defined by 8 points for the detected vessel.

\subsection{Vessel Detection}
\label{sec:ship_detection}
Many object detection models can be effectively applied to vessel detection using their "off-the-shelf" weights, thanks to datasets like MS COCO \cite{lin2014microsoft}, which includes a \textit{boat} class. In this context, we denote the output of the model as $BB_{pred}$, which corresponds to the predicted vessel locations in image space. To demonstrate that our technique does not rely on specialised models, we conducted experiments (see \cref{sec:object_detection_results}) using generalised models. 

For vessel detection, we selected YOLOX-X \cite{ge2021yolox}, a variant of the \ac{yolo} family \cite{redmon2016you}, which builds upon the \ac{yolo}v3 architecture \cite{redmon2018yolov3}. The YOLOX model family offers several sizes: YOLOX-X, YOLOX-L, YOLOX-M, and YOLOX-S, which differ primarily in model complexity, size, and computational requirements. The smaller YOLOX-S and YOLOX-M variants are optimised for scenarios with limited computational resources, providing faster inference at the expense of some accuracy. In contrast, the larger YOLOX-L and YOLOX-X models deliver higher accuracy but require more computational power and have longer inference times.

\ac{yolo} models are widely recognised for their real-time object detection capabilities, thanks to their efficient single-stage detection approach that balances speed and accuracy \cite{diwan2023object}. Our goal is to establish a well-balanced baseline solution that can be further extended in future work.

\subsection{Coordinate Transformation}
Transforming world coordinates to image coordinates is required to align the georeferenced \ac{ais} data with object detections in image space. For this, we will compare homography based approach as used in related work \cite{solano2021detection, carrillo2022ship, gulsoylu2024image}, and a full pinhole camera model, whose parameters are estimated with \ac{pnp}. Both of these methods require a set of keypoints where the coordinates are known in both image space and in real world coordinate system.

\subsubsection{Keypoint Selection}
We manually identified and annotated a set of points that are easily recognisable both in the real world and in the image. These points include distinct features such as the corners of buildings. For each of these points, we recorded their real-world coordinates and their corresponding coordinates in the image. 

Homography assumes that all keypoints points lie on the same plane, making it useful for applications such as image stitching \cite{nie2020view}. For our case, we can consider the water surface as a planar surface. However, defining keypoints is challenging due to the difficulty in distinguishing points on the water surface. Selecting keypoints on land is easier, but land surfaces are not coplanar with the water surface. Even without considering tidal effects, it is challenging to select keypoints where water intersects with a distinct artificial structure, such as a pier. We defined our points using artificially built structures, not all of these points are coplanar with the water surface.

To make our corresponding points suitable for homography, we used two approaches to make our real world keypoints coplanar on an interim plane. In one, we determined the tangential plane that touches the water surface that has the normal of the earth's surface at this point. In another approach, we applied \ac{pca} on our set of 3D world corresponding points. \ac{pca} finds the axes along which the variance of the set is represented strongest. This allows us to choose the two axes that represent our 3D points in the best way possible, meaning losing the least information by excluding the least important principal component. With both approaches, we can arrive at two vectors in our 3D world space, that span an interim plane onto which we can project our corresponding points.

\subsubsection{Homography}
%Maybe we dont even need this paragraph? Or a shortened version.
%Homography is a technique used in computer vision to establish a relationship between two images of the same planar surface. It enables the transformation of coordinates from one image to another using a homography matrix \cite{vincent2001detecting}. 
Homography is a technique that enables the transformation of coordinates from one plane to another using a homography matrix \cite{vincent2001detecting}. 

$\mathbf{A}$ is the matrix with which we project our 3D points onto the interim plane. It is defined as
\begin{equation}
\mathbf{A} = \begin{bmatrix} \mathbf{v}_1 & \mathbf{v}_2 \end{bmatrix},
\end{equation}
where $\mathbf{v}_1$ and $\mathbf{v}_2$ are 3D vectors, either two orthogonal vectors of the tangential plane, or the two most important principal components.

We project our 3D keypoints $\mathbf{k}_{\text{3d}}$ onto the plane, calling them $k_{2d}$ using the following transformation:
\begin{equation}
\mathbf{k}_{\text{2d}} = \mathbf{A}^T \mathbf{k}_{\text{3d}}.
\end{equation}

The Homography matrix, $\mathbf{H}$, can be approximated by minimising the reprojection error when mapping the real world coordinates on the interim plane to known image coordinates. Once $\mathbf{H}$ is obtained, we can transform real world coordinates into the image.

If $\mathbf{x}_1 = (u, v, 1)$ is a pixel location and $\mathbf{x}_2 = (x, y, 1)$ is a  coordinate in our interim plane in the real world, we can transform between world and image by finding $\mathbf{H}$ in
\begin{equation}
	\mathbf{x}_1 = \mathbf{H} \mathbf{x}_2.
\end{equation}
This mapping with a homography matrix between two planes skips the camera projection, avoiding the need to determine the camera's intrinsic and extrinsic matrix.
        
\subsubsection{Perspective-n-Point}
The \acf{pnp} algorithm is useful for estimating the pose (position and orientation) of a camera given a set of 3D points in world coordinates and their corresponding 2D projections in an image \cite{lu2018review}. This method requires a more complex model of the relationship between real-world and image space, compared to homography. Here, we do not have to project our keypoints onto a interim plane in the real world, we can directly use the points spanning all three dimensions. We used a pinhole camera model \cite{marchand2015pose} defined as follows: 
\begin{equation}
	\overline{\mathbf{x}} = \mathbf{K} \; \mathbf{\Pi}  \; {}^c\mathbf{T}_w  \; {}^w\mathbf{X}.
    \label{eq:camera_projection}
\end{equation}

A point, ${}^w\mathbf{X}$, in the real world is first projected into a frame of reference relative to the camera, using the extrinsic parameters of the camera, ${}^c\mathbf{T}_w$. This matrix contains the 3D position ${}^c\mathbf{t}_w$ of the camera, as well as its rotation in the world ${}^c\mathbf{R}_w$. $\mathbf{\Pi}$ is the projection matrix of a perspective projection, reducing the dimensionality by projecting onto a plane. $\mathbf{K}$ is the intrinsic camera matrix containing its field of view, aspect ratio, etc. The result is an image space coordinate $\overline{\mathbf{x}}$ \cite{lindashapiro}.

The camera's intrinsic matrix $\mathbf{K}$ is approximated for the specific camera model using chequerboard calibration. Given $\mathbf{K}$ and a set of $n$ point correspondences in 2D to 3D, the extrinsic matrix ${}^cT_w$ is approximated with \ac{pnp}.

With the established capability to perform transformations between world coordinates and image space, we can now project \ac{ais}-based data, such as vessel’s location, onto the corresponding image. 

\subsection{Defining water-level plane segment}
The water-level plane segment is defined by four 3D corner points on the water surface in world coordinates, and serves as the geometric base for constructing the full 3D bounding box for 6D pose estimation.

\subsubsection{Data in AIS}
\ac{ais} data is transmitted as a collection of standardised message types. The most relevant for vessel localisation and tracking are Position Reports, which carry dynamic information like position, speed, and course. Furthermore there are Static and Voyage Related Data messages, including vessel identity, type, dimensions, and voyage details. 

\ac{ais} emitters are categorised based on the class of the transceiver: Class A and Class B \cite{norris2006automatic}. Class A devices are mandatory for larger commercial vessels and transmit at higher power and more frequent intervals, providing more detailed navigational data. In contrast, Class B devices are designed for smaller vessels, transmit less frequently, and at lower power, carrying a reduced set of data fields compared to Class A.  

In our work, we focussed on Class A \ac{ais} messages.
By collecting these types of \ac{ais} messages, we can aggregate them by vessel identity to create a initial pose per vessel.

\subsubsection{Using data from AIS to create a initial 6D pose}

\begin{figure}[]
	\centering
    %TODO: a_3d is used twice!!!
		\includegraphics[width=\linewidth,trim=.4cm 3cm 3.5cm 3.5cm, clip]{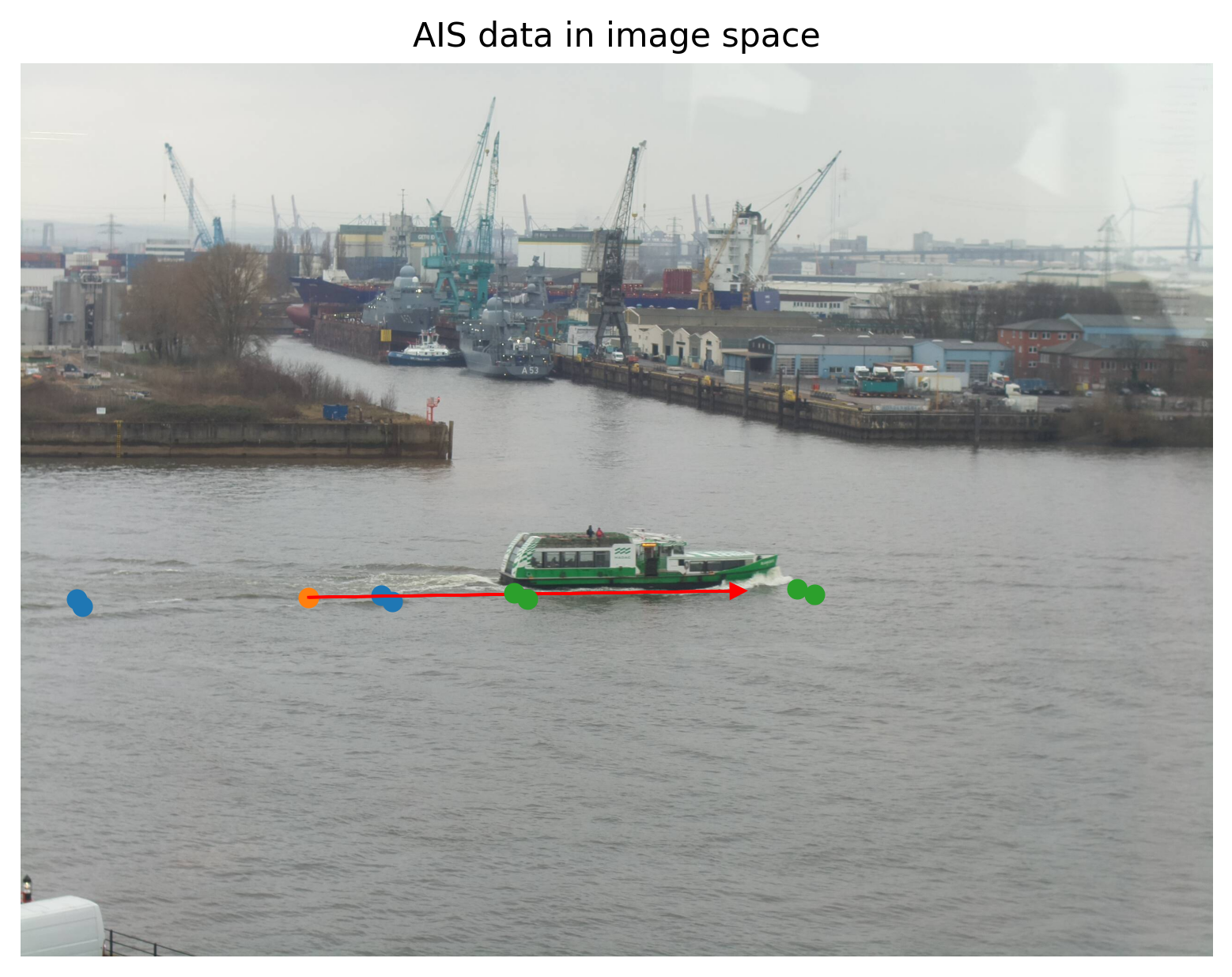}
		\caption{\ac{ais}-based vessel location, \( \mathbf{a}_{\text{3d}}\) (orange) and water-level plane segment corners \( c_{\text{3d}} \) (blue). Dead reckoning vector, \( \mathbf{dr} \) (red) and corrected water-level plane segment corners \( \mathbf{c}_{\text{3d}} \) (green) after dead reckoning. The older position in \ac{ais} does not match up with the image, dead reckoning improves this, but the inaccuracy of the \ac{ais} pose still leaves a small mismatch.}
		\label{fig:AisProjectionResult}
\end{figure}

The location data transmitted in \ac{ais} by the vessels is obtained  via \ac{gps}.
\ac{gps} data is based on EPSG:4979 \cite{epsg4979}, a geographic coordinate system that represents locations on the Earth's surface with angles (\textit{longitude} and \textit{latitude}) on a spherical Earth model. This type of representation is not fitting for the 6D pose estimation problem. Therefore, we converted the geographic coordinate (EPSG:4979) into a geocentric coordinate system EPSG:4978 \cite{epsg4978} that represents locations using three orthogonal axes. This makes the \ac{ais} position compatible to our camera model.

With the aforementioned information from \ac{ais} we create a right handed object coordinate system $v$ for each vessel in our three dimensional world space, representing the vessels position and orientation in it. The axis that points forward is $\mathbf{x}_v$. It lies on the tangential plane of the Earth at the point of the \ac{ais} antenna. We assumed roll and pitch to be zero, since vessels operate close to upright pose. Thus its direction is determined solely by the heading of the vessel.  Likewise, $\mathbf{y}_v$ is the orthogonal axis, that points to the vessels port side. $\mathbf{z}_v$ is the "up" vector, that is the tangential plane's normal. All axes converge at the point determined by the \ac{ais} antenna location. The vectors used as the object system axes are normalised to a metre length. The origin and orientation of $v$ represent the  pose of the vessel, based on \ac{ais} data.

Together with the vessel dimensions from the static \ac{ais} report, we define the corners of the water-level plane segment, 3D points in our world space, as follows:
\begin{align}
	\mathbf{c1}_{\text{3d}} &= \mathbf{a}_{\text{3d}} + (A \cdot \mathbf{x}_v) + (C \cdot \mathbf{y}_v) \\
	\mathbf{c2}_{\text{3d}} &= \mathbf{a}_{\text{3d}} + (A \cdot \mathbf{x}_v) + (D \cdot (-\mathbf{y}_v)) \\
	\mathbf{c3}_{\text{3d}} &= \mathbf{a}_{\text{3d}} + (B \cdot (-\mathbf{x}_v)) + (D \cdot (-\mathbf{y}_v)) \\
	\mathbf{c4}_{\text{3d}} &= \mathbf{a}_{\text{3d}} + (B \cdot (-\mathbf{x}_v)) + (C \cdot \mathbf{y}_v)
\end{align}
where A is the distance to bow, B is to stern, C is the distance to port and D is the distance to starboard from the \ac{ais} antenna location $\mathbf{a}_{\text{3d}}$. (see blue points in \cref{fig:AisProjectionResult}).
Having established the corners of the water-level plane segment using the vessel's dimensions and location, it is important to consider the update frequency of the \ac{ais} messages. At the moment of image acquisition, the most recent \ac{ais} message from the corresponding vessel may be up to ten seconds old, resulting in outdated location information. We create a dead reckoning vector, $\mathbf{dr}$, to mitigate the temporal inaccuracy in vessel location data caused by the update frequency of \ac{ais} messages (see the red arrow in \cref{fig:AisProjectionResult}). By scaling the forward vector $\mathbf{x}_v$ with the vessel's speed and the age of the \ac{ais} message at the time of image acquisition, we update the location of the water-level plane segment (see orange points in \cref{fig:AisProjectionResult}). 
% Our \ac{dr} technique assumes the vessel maintains its last known direction and speed.

\subsection{Matching AIS data with 2D detection results}
At this stage, the vessels are detected using an object detection model and their \ac{ais} based location is projected onto the images as a water-level plane segment. However, we still need to compensate for the inaccuracies in \ac{ais} location data and derive the air draught information to create accurate 3D bounding boxes for the detected vessels. As the next step, we want to fuse the two data sources by first associating the 2D bounding box, $BB_{pred}$, with the water-level plane segment.

We formulate the task of matching \ac{ais} messages to detected vessels as a bipartite graph matching problem \cite{serratosa2014fast}. One set of nodes are the detection bounding boxes of our object detector, $BB_{pred}$. The other set of nodes is derived from the \ac{ais} information in image space $WP_{ais}$. These two sets of nodes are fully connected in the graph, each edge being assigned the result of a weight function described in \cref{eq:theta_cases}, representing a matching cost. This allows tuning matching performance by adjusting the weight function.

The object prediction nodes contain the information about the 2D bounding box and the confidence score of the network. $BB_{pred}$ is such a prediction, defined as 
\begin{equation}
BB_{pred} =  [x_1, y_1, x_2, y_2, \text{score}]
\end{equation}
where ($x_1$,~$y_1$) and ($x_2$,~$y_2$) are the two corners spanning the bounding box and score is the detection confidence.

We project our set of water-level plane segment points from world space $c_{\text{3d}}$ into image space $c_{\text{img}}$ using the camera model determined via \ac{pnp}.
The water-level plane segment is projected onto the image space as a convex quadrilateral. For these four points $c_{\text{img}}$, we create the minimal enclosing rectangle, $WP_{ais}$. This is so it has the same shape as the rectangular object prediction:
\begin{equation}
    WP_{ais} = [\min_x(c_{\text{img}}), \min_y(c_{\text{img}}), \max_x(c_{\text{img}}), \max_y(c_{\text{img}})]
\end{equation}
where $\min_x(c_{\text{img}})$ is the minimal x-coordinate of all water-level plane segment corners in the collection $c_{\text{img}}$, and the other min/max functions are defined accordingly. 

The matching cost is calculated based on the distance between the predicted 2D bounding box, $BB_{pred}$, and the water-level plane segment, $WP_{ais}$, as well as the difference in their widths. These measures are taken in image space. A correct match between $WP_{ais}$ and $BB_{pred}$ is expected to show spatial proximity, with the width of $BB_{pred}$ being approximately proportional to the $WP_{ais}$.

With the width \( w \) and height \( h \) of the image, the edge weight \( \theta \) is defined as:

\begin{equation}
	\Delta \text{cx} = \frac{\left| \frac{x_1 + x_2}{2} - \frac{\min_x(c_{\text{img}}) + \max_x(c_{\text{img}})}{2} \right|}{w}
    \label{eq:delta_centre_x}
\end{equation}

\begin{equation}
	\Delta \text{bottom} = \frac{\left|y_2 - \max_y(c_{\text{img}})\right|}{h}
    \label{eq:delta_bottom}
\end{equation}

\begin{equation}
	\Delta \text{width} = \frac{\left|(x_2 - x_1) - (\max_x(c_{\text{img}})-\min_x(c_{\text{img}}))\right|}{w}
    \label{eq:delta_width}
\end{equation}

{\small
\begin{equation}
    \theta(BB_{pred_{i}}, WP_{ais_{j}}) =
    \begin{cases}
        \dfrac{2\Delta \text{cx} + \Delta \text{bottom} + \Delta \text{width}}{\text{score}},\\
        \quad \text{if vessel not on border} \\
        \\
        \dfrac{2\Delta \text{cx} + \Delta \text{bottom}}{\text{score}},\\
        \quad \text{if vessel on border}.
    \end{cases}
    \label{eq:theta_cases}
\end{equation}
}

We compute the distance between $BB_{pred}$ and $WP_{ais}$ as the sum of the distance between the centres of their x-coordinates (\cref{eq:delta_centre_x}), and the distance between their lower edges along $y$ axis (\cref{eq:delta_bottom}). 
%@Emre: Added explanation sentence
This special treatment is necessary because we cannot meaningfully compare the upper edge of the prediction bounding box with the water-level plane segment, as the latter \ac{ais} information source contains no vessel height. We assign greater weight to the centre factor, as tide-induced variations can introduce inaccuracies in the vertical placement of the projected \ac{ais} bounding box, which are not accounted for in the projection.

The difference in bounding box width (\cref{eq:delta_width}) is only considered if the prediction bounding box does not lie on the image borders (\cref{eq:theta_cases}). When a prediction is on the image border, it indicates that the vessel is not fully visible. In such cases, comparing the width of the clipped prediction with the width of the non-clipped \ac{ais} projection would be misleading.
%@Emre maybe remove the remaining paragraph?
The missing $\Delta \text{width}$ factor favours matches on the image border, due to lowering the weight \( \theta \). At the same time the $\Delta \text{cx}$ component is likely higher in those cases, as $BB_{pred}$ is constrained to the image size while $WP_{ais}$ can protrude, leading to an increase in the $x$ centre difference. These factors oppose each other, so that we cannot find evidence for this design choice degrading our matching performance.

The score acts as a scaling factor, favouring high-confidence predictions while still allowing lower-confidence predictions with better spatial compatibility to be matched.

The graph method efficiently matches 2D bounding box predictions from the object detection model with \ac{ais}-derived boxes by finding the minimum weight matching. Due to its fully connected nature, there will always be a match possible. To improve accuracy, we set a maximum weight threshold to filter out irrelevant matches. This refinement allows the system to return "no match" for significant spatial discrepancies, size differences, or low confidence scores, ensuring only the most reliable matches are considered. 
% A sample match is visualised in \cref{fig:correction_and_3D_bbox} (top image).

% \begin{figure}[]
% 	\centering
% 		\includegraphics[trim={0 50px 0 50px}, clip, width=1\linewidth]{imgs/methodology_steps/matches.png}
% 		\caption{A sample image of matched 2D bounding box $BB_{pred}$ (blue rectangle) and the water-level plane $WP_{ais}$ (blue points). The text is the anonymised name of the vessel. Even after dead reckoning correction, the water-level plane may not exactly match with the predicted 2D bounding box due to spatial inaccuracies of \ac{ais}.}
% 		\label{fig:bb_matches}
% \end{figure}

% \begin{figure}
%     \centering
%     \subfloat{
%         \includegraphics[trim={0 50px 0 50px}, clip, width=0.95\linewidth]{imgs/methodology_steps/matchingMatchNoName.png}
%     }
%     \vspace{-7pt}
%     \subfloat{
%         \includegraphics[trim={0 50px 0 50px}, clip, width=0.95\linewidth]{imgs/methodology_steps/finalPoses.png}
%     }
%     \caption{Top, a sample image of matched 2D bounding box $BB_{pred}$ (blue rectangle) and the water-level plane segment $WP_{ais}$ (blue points). Even after dead reckoning correction, the water-level plane segment does not exactly match with the object detection 2D bounding box due to inaccuracies in \ac{ais}. Bottom, correcting the water-level plane segment corners with the detection bounding box and determining the vessel height allows the creation of a 3D bounding box that encloses the vessel.}
%     \label{fig:correction_and_3D_bbox}
% \end{figure}

\subsection{Correcting AIS data using visual cues}
At this stage we have, for a single vessel instance, a match of a 2D bounding box in image space and a corresponding water-level plane segment derived from \ac{ais} data. These two can deviate spatially; visual cues can help reducing this spatial deviation. For the rotation we rely on the \ac{ais} data, but for the location, we can use the 2D bounding box in image space as a constraint for the position of the water-level plane segment. This allows us to correct both temporal and spatial-accuracy errors in \ac{ais} data. We refine the \ac{ais}-based location prior by applying a correction vector $\Delta \mathbf{v}_{\text{3d}}$ to all points of our water-level plane segment, yielding the bottom four corners of our final 3D bounding box:

\begin{equation}
    \mathbf{b}_{i\text{3d}} = \mathbf{c}_{i\text{3d}} + \Delta \mathbf{v}_{\text{3d}}, \quad \text{for } i = 1, 2, 3, 4.
\end{equation}

$\Delta \mathbf{v}_{\text{3d}}$ is determined with the help of our camera model (\cref{eq:camera_projection}):

Since the transformation between image space and real-world space has been estimated, the edges of the 2D bounding box in image space can be brought into the real world by reversing the perspective projection (\cref{eq:camera_projection}). Each edge of the bounding box is transformed into a plane representing a constraint, and all four planes converge at the camera's location in the real world, $\mathbf{C}$. This location is determined by the extrinsic camera matrix, a parameter determined by \ac{pnp}:
\begin{equation}
	\mathbf{C} = -{}^c\mathbf{R}_w^T~{}^c{\mathbf{t}}_w
\end{equation}

We wrap the projection reversal of the camera model (\cref{eq:camera_projection}) in a function named $pld$ (projection line direction) so that
\begin{equation}
 L(\lambda) = \mathbf{C} + \lambda \cdot \text{pld}(\mathbf{p}_{\text{img}})
\end{equation}
$L$ is a line in world space on which all points are projected to the same pixel $p_{\text{img}}$ and $\lambda$ is the distance of a point in the line from the camera.

The correction vector, $\Delta \mathbf{v}_{\text{3d}}$, is determined with spatial constraints using three camera model based planes ($\alpha$,~$\beta$,~$\gamma$) corresponding to left, right and bottom edges of the 2D bounding box. 
The four corners of the 2D bounding box in image space $BB_{pred}$ are referred to as: $\mathbf{p_1} = (x_1, y_1), \quad \mathbf{p_2} = (x_1, y_2), \quad \mathbf{p_3} = (x_2, y_1), \quad \mathbf{p_4} = (x_2, y_2)$.

The plane $\alpha$ is established to be positioned such that, when viewed from the camera's perspective, the entire vessel lies to the right of it.

The plane $\alpha$ with its normal vector $\mathbf{n}_{\alpha}$ are defined as:
\begin{equation}
	\mathbf{n}_{\alpha} = \mathbf{z}_v \times \text{pld}(\mathbf{p_2}_{\text{img}})
\end{equation}
\begin{equation}
	\alpha : \{ \mathbf{X} \mid (\mathbf{X} - \mathbf{C}) \cdot \mathbf{n}_{\alpha} = 0 \}
\end{equation}
$\alpha$ is orthogonal to the water surface and the projection line of $\mathbf{p_2}_{\text{img}}$ lies in $\alpha$.

For each corner $\mathbf{c}_{\text{3d}}^{(n)} \in c_{3d}$ of the water-level plane segment in world space we calculate the distance to the plane $\alpha$.
\begin{equation}
	d_{\alpha}(\mathbf{c}_{\text{3d}}^{(n)}) = \frac{(\mathbf{c}_{\text{3d}}^{(n)} - \mathbf{C}) \cdot \mathbf{n}_{\alpha}}{\|\mathbf{n}_{\alpha}\|}
    \label{eq:alpha_distance_function}
\end{equation}
In this calculation, the distance can be negative, depending on which "side" of the plane $\mathbf{c}_{\text{3d}}^{(n)}$ is.

Similarly, we define a plane, $\beta$, for the right edge and another plane, $\gamma$ for the bottom edge. Then the related distance measures $d_{\beta}(\mathbf{c}_{\text{3d}}^{(n)})$ and $d_{\gamma}(\mathbf{c}_{\text{3d}}^{(n)})$ are calculated for $\beta$ and $\gamma$, respectively. 

Each plane can yield a proposal for $\Delta \mathbf{v}_{\text{3d}}$ that makes the water-level plane segment compliant with its constraint. We define $\Delta \mathbf{v}_{3d}$ by averaging the correction proposals determined by the three planes. The goal is to shift the water-level plane segment in the space between the three planes ($\alpha$,~$\beta$,~$\gamma$). The top case is the default one, using all correction planes, the other special cases exclude the correction factor of a plane if it protrudes the image border: 
\begin{align}
	\Delta \mathbf{v}_{\text{3d}} &= 
	\begin{cases}
		d_{\gamma}^{(k)} \mathbf{n}_{\gamma} + \frac{d_{\alpha}^{(i)} \mathbf{n}_{\alpha}}{2} + \frac{d_{\beta}^{(j)} \mathbf{n}_{\beta}}{2}, & \\
		\quad \text{if }BB_{pred} \text{ in image} \\[10pt]
		d_{\gamma}^{(k)} \mathbf{n}_{\gamma} + d_{\alpha}^{(i)} \mathbf{n}_{\alpha}, & \\
		\quad \text{if }BB_{pred} \text{ protrudes right}, \\[10pt]
		d_{\gamma}^{(k)} \mathbf{n}_{\gamma} + d_{\beta}^{(j)} \mathbf{n}_{\beta}, & \\
		\quad \text{if }BB_{pred} \text{ protrudes left},\\[10pt]
		d_{\gamma}^{(k)} \mathbf{n}_{\gamma}, & \\
		\quad \text{if }BB_{pred} \text{ protrudes both sides.}
	\end{cases}
\end{align}

% The proposals of $\alpha$ and $\beta$ will conflict if the size of predicted 2D bounding box and the water-level plane do not match up. For example, if the water-level plane is wider than the predicted 2D bounding box, $\alpha$ and $\beta$ will push $a_{\text{3d}}$ in different directions, cancelling each other out.

% If the boxes are of the same size, and the \ac{ais} bounding box is left of the prediction bounding box, the correction will be applied via both the $\alpha$ and the $\beta$ correction, potentially ending up over-correcting to the right.
%TODO FH: A weakpoint here is: There should be a case for also excluding the gamma restraint when bb_pred touches the lower edge of the image. In our dataset this never happened, thats why I only thought about it now.

Correction in the image's depth is always applied via $\gamma$, side-wise correction is applied conditionally.
In the first case, an average of the $\alpha$ and $\beta$ corrections is applied when the object prediction bounding box does not touch the image edges. The other cases ignore the correction planes proposal if the bounding box edge they are derived from is caused by the image border.
%@Emre: Shortened explanation that is now evident in equation by renaiming conditions to human readable explanations.
%In the second case, the right edge is discarded for correction purposes when it is on the image edge, and only the $\alpha$ plane is considered for correction. Similarly, the $\beta$ plane is used for correction when the left edge of the predicted 2D bounding box is on the image edge. No side-shift is applied to the water-level plane segment when the 2D bounding box touches both image edges. With $\Delta \mathbf{v}_{3d}$ determined, we can correct the water-level plane segment, minimising the error caused by temporal and spatial inaccuracy of the \ac{ais} data.

\subsection{Creating the 3D bounding box}
\label{sec:datafusionAisPred}
Given the spatially corrected water-level plane segment, we can now complete the 3D bounding box, \( b_{\text{3d}} \), that fully encompasses the vessel in 3D space. To substitute the lack of air draught information in the \ac{ais} data, we calculate an estimated height of the vessel, $h_v$, by shifting the water-level plane segment, $\mathbf{c}_{\text{3d}}$ after correction, upwards, along the water surface's normal vector with $\Delta \mathbf{h}_{\text{3d}} = h_v \cdot \mathbf{z}_v$, aligning it with the top edge of the 2D bounding box.

We create a forth restriction plane $\delta$ using the upper edge of the 2D bounding box:
\begin{align}
	\mathbf{n}_{\delta} &= \text{pld}(\mathbf{p1}_{\text{img}}) \times \text{pld}(\mathbf{p3}_{\text{img}}) \\
	\delta &: \{ \mathbf{X} \mid (\mathbf{X} - \mathbf{C}) \cdot \mathbf{n}_{\delta} = 0 \}
\end{align}

The distance function is defined similar to \cref{eq:alpha_distance_function}, but taking the correction vector into account:
\begin{equation}
	d_{\delta}(\mathbf{c}_{\text{3d}}^{(n)}) = \frac{(\mathbf{c}_{\text{3d}}^{(n)} + \Delta \mathbf{v}_{\text{3d}} - \mathbf{C}) \cdot \mathbf{n}_{\delta}}{\|\mathbf{n}_{\delta}\|}
\end{equation}

%Check calculations-> delta normal isn't exactly the water normal
where \( d_{\delta}^{(l)} \) is the distance to move our water-level plane segment upwards, until it touches the restriction plane $\gamma$. We thus set $h=d_{\delta}^{(l)}$.

We complete the 3D bounding box creating the top four corners by translating the corrected water-level plane segment \( c_{\text{3d}} \) with the offset vector \( \Delta \mathbf{h}_{\text{3d}} \) as follows: 
\begin{equation}
    \mathbf{b}_{i\text{3d}} = \mathbf{c}_{(i - 4)\text{3d}} + \Delta \mathbf{v}_{\text{3d}} + \Delta \mathbf{h}_{\text{3d}}, \quad \text{for } i = 5, 6, 7, 8.    
\end{equation}

\begin{figure*}[]
	\centering
	\subfloat{\includegraphics[width=0.33\textwidth, trim=1.5cm .5cm 1.5cm 2.5cm, clip]{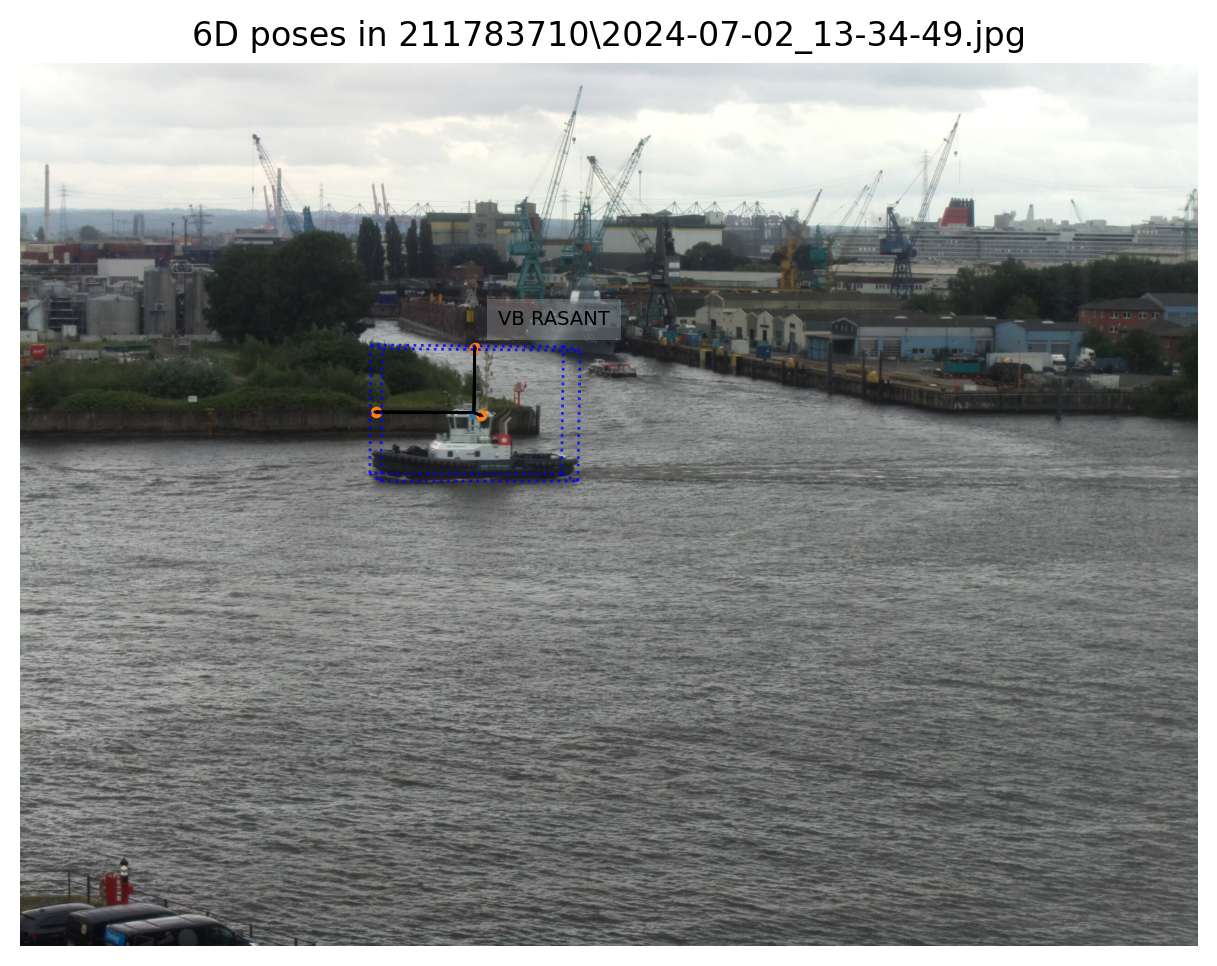}} \hfill
	\subfloat{\includegraphics[width=0.33\textwidth, trim=1.5cm .5cm 1.5cm 2.5cm, clip]{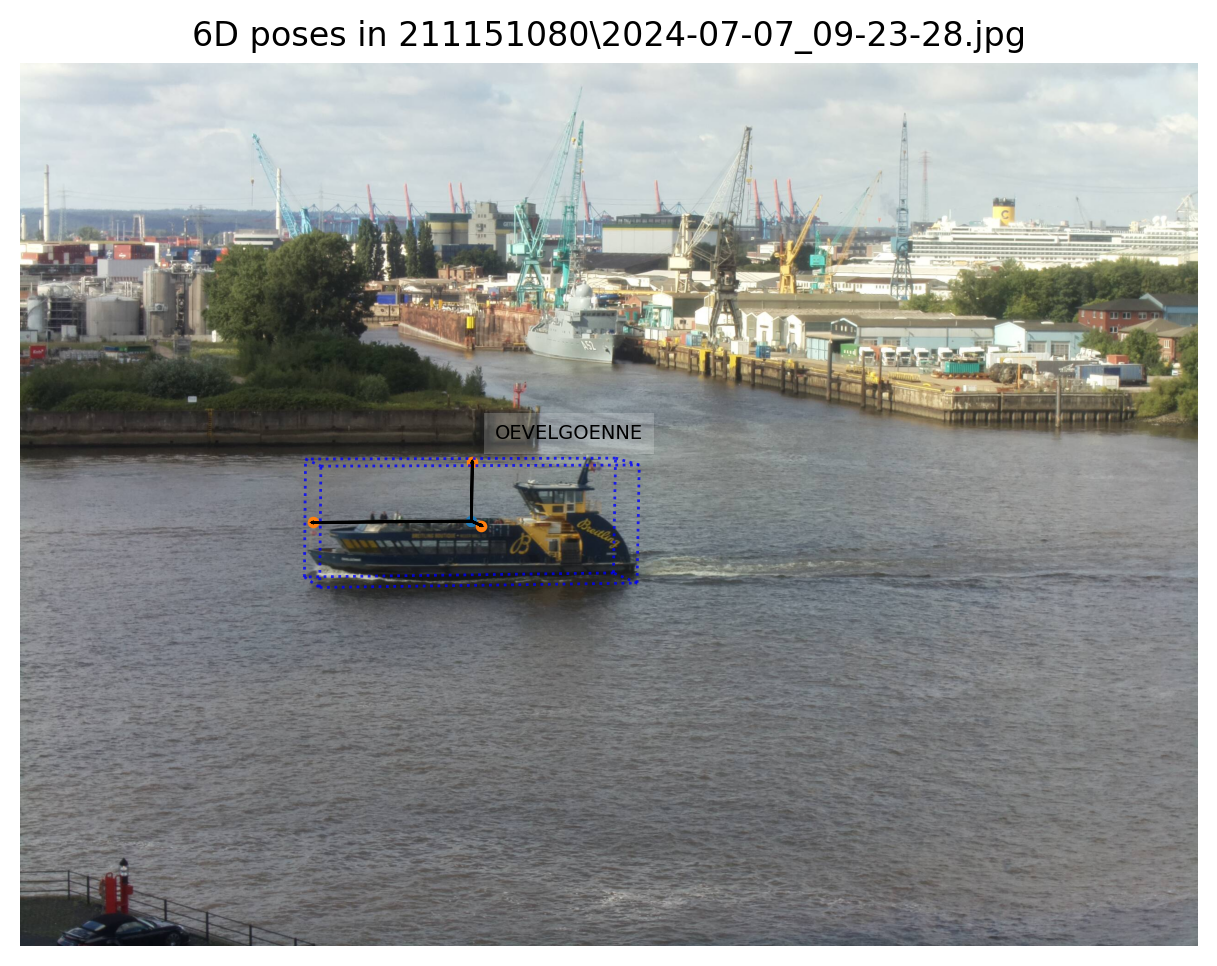}} \hfill
	\subfloat{\includegraphics[width=0.33\textwidth, trim=1.5cm .5cm 1.5cm 2.5cm, clip]{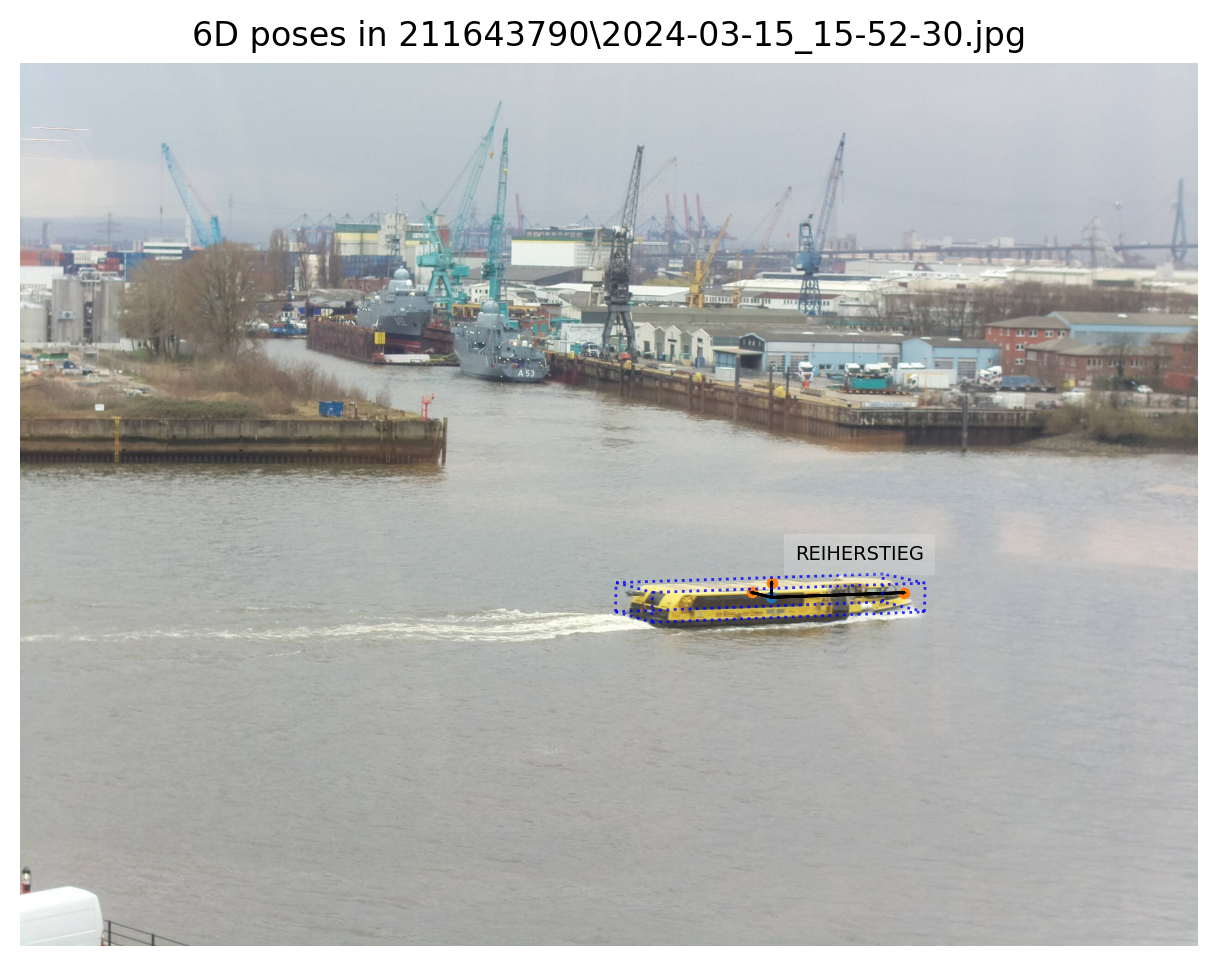}} \hfill
    
    \vspace{-8pt}
    
	\subfloat{\includegraphics[width=0.33\textwidth, trim=1.5cm .5cm 1.5cm 2.5cm, clip]{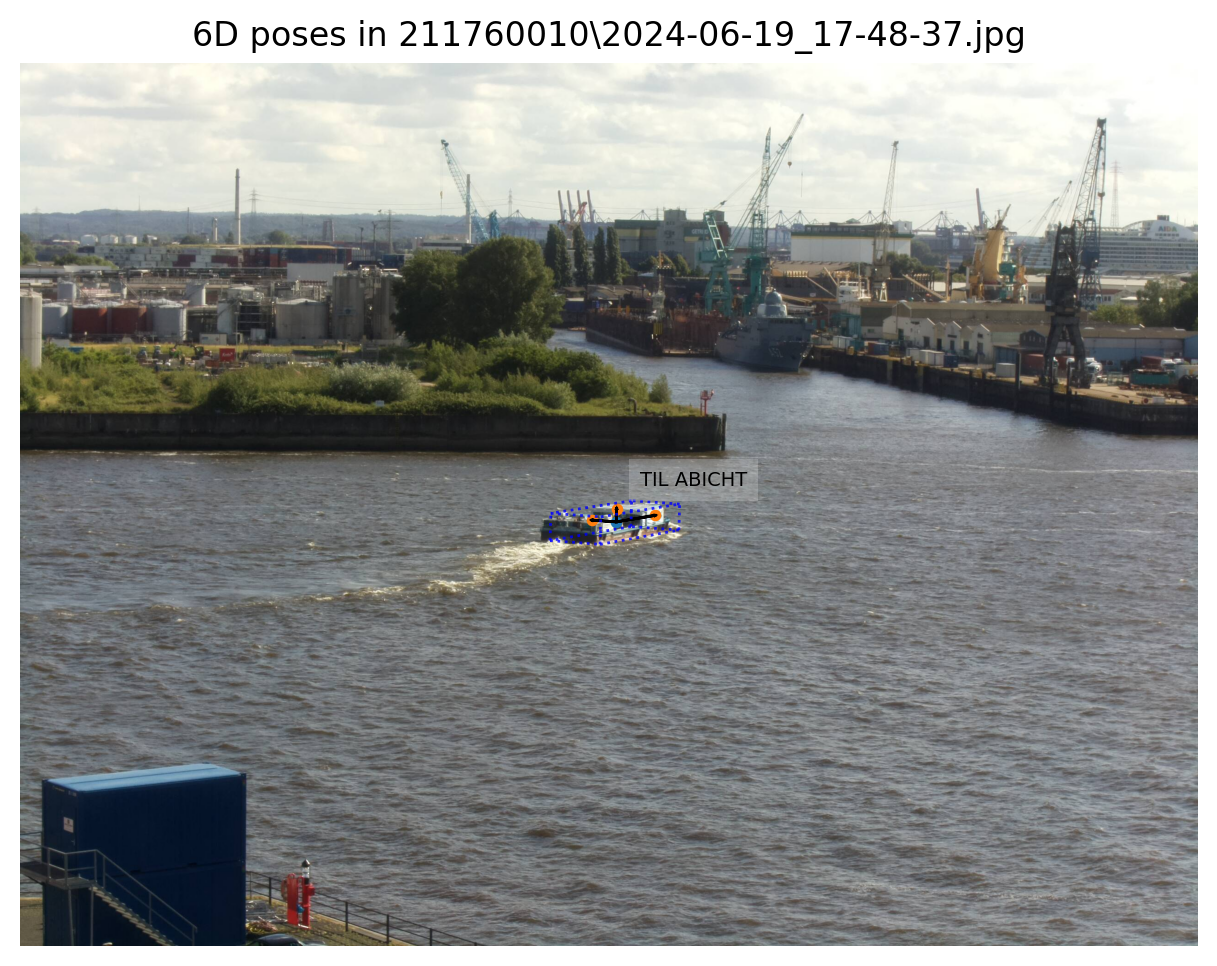}} \hfill
	\subfloat{\includegraphics[width=0.33\textwidth, trim=1.5cm .5cm 1.5cm 2.5cm, clip]{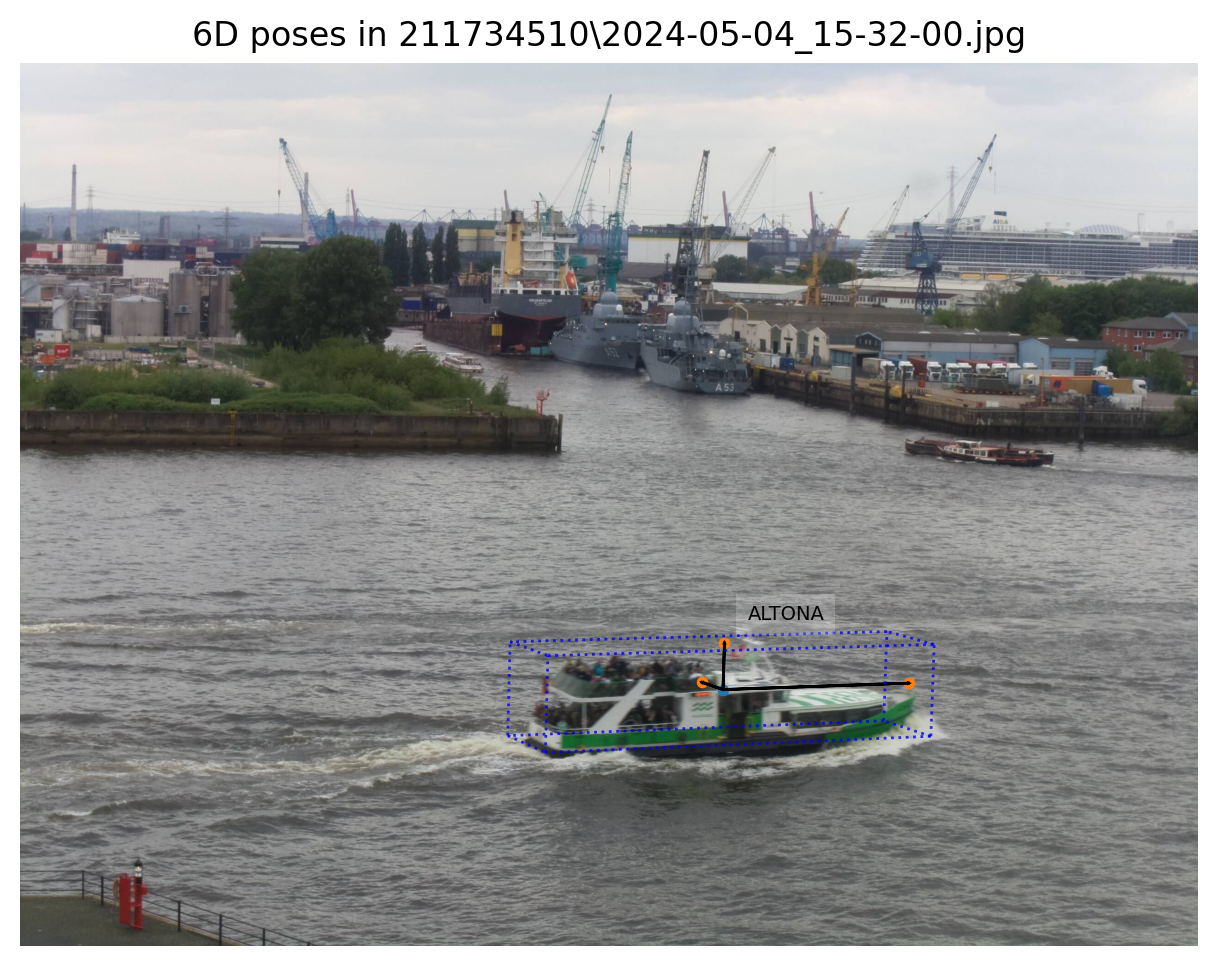}} \hfill
	\subfloat{\includegraphics[width=0.33\textwidth, trim=1.5cm .5cm 1.5cm 2.5cm, clip]{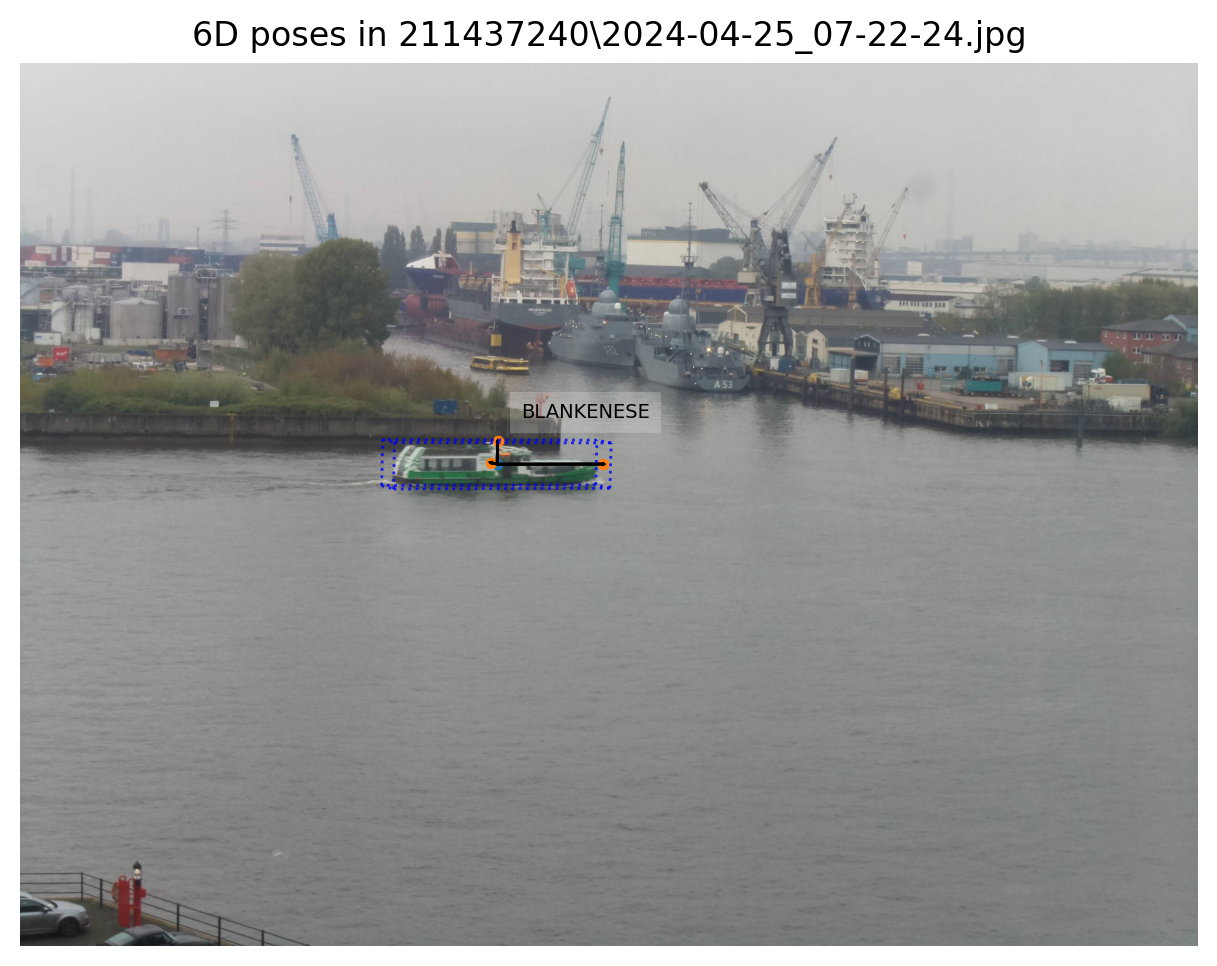}} \hfill
	% \subfloat{\includegraphics[width=0.33\textwidth, trim=1.5cm .5cm 1.5cm 2.5cm, clip]{imgs/6D_dataset_result_samples/6.png}} \hfill
    % \vspace{-8pt}
    
	% \subfloat{\includegraphics[width=0.33\textwidth, trim=1.5cm .5cm 1.5cm 2.5cm, clip]{imgs/6D_dataset_result_samples/7.png}} \hfill
	% \subfloat{\includegraphics[width=0.33\textwidth, trim=1.5cm .5cm 1.5cm 2.5cm, clip]{imgs/6D_dataset_result_samples/9.png}} \hfill

	% \caption{Nine cropped samples from the resulting dataset with their pose annotated. This sample shows how most vessels drive straight by the camera, leading to a skewed distribution of headings.}
    \caption{Sample images from our dataset with pose annotations visualised.}
	\label{fig:PoseDatasetSamples}
\end{figure*}

The bounding box created this way now spans the vessel in 3D space.
% \cref{fig:correction_and_3D_bbox} shows a initial match of a 2D bounding box and a water-level plane segment (top) as well as the final 3D bounding box result of the described correction (bottom image).
We can calculate the vessel's centroid by averaging the corners, and the vessel's rotation is given by its object axes $\mathbf{x}_v, \mathbf{y}_v, \mathbf{z}_v$. Sample images with 6D pose annotations are visualised in \cref{fig:PoseDatasetSamples}. 

\section{Boats on Nordelbe Kehrwieder Dataset}
% \begin{itemize}
%     \item Dataset fOr Pose Estimation on Norderelbe (DOPEN)
%     \item POSe Estimation on Norderelbe Dataset (POSEND)
%     \item NEH6D
% \end{itemize}
The \ac{bonk-pose} dataset we publish consists of 3753 images featuring a total of 3829 vessels, each annotated with their respective pose. The annotations include the vessels 3D centroid, its 3D dimension and rotation in 3D space. These 6D pose annotations are created by our automatic approach based on fusing the vessels' dimensions, heading and location from \ac{ais} messages with visual cues.

We initially collected 4176 monocular RGB images around the River Elbe, and relevant \ac{ais} messages. Applying our approach, we were not able to find \ac{ais} to vessel detection matches in all of them. We will look at the reasons for this in the \cref{sec:evaluation}. A first comparison between vessel detections and the presence of vessel emitting \ac{ais} data already reveals discrepancies. The object detector often identifies more vessels in the images than are indicated by the \ac{ais} data. This mismatch may result from vessels not transmitting \ac{ais}, incomplete \ac{ais} data, or inaccuracies in the object detection model. We excluded those images from the final dataset, in which we were not able to find at least a single match.

The final dataset includes a diverse set of vessels: The occurrence distribution is skewed towards the public transport ferries and  harbour tour vessels that regularly pass by the field of view. On the other hand, nearly 70 vessels are captured only once. Analysing the vessel sizes, we observed that most of them are relatively small, with lengths under 100 meters and widths less than five meters. However, the dataset also includes some larger vessels, providing a representation of different vessel types. Vessel locations are mostly centred in the middle of the field of view. The vessel centroids are 150 to 400 metres away from the camera position. Since the images are taken in an inland waterway, the distribution of the vessel rotations is less diverse, most vessels captured side-on. 

\begin{figure*}
    \centering
    \subfloat{
        \includegraphics[width=0.48\textwidth, trim=0cm 7cm 0cm 1cm, clip]{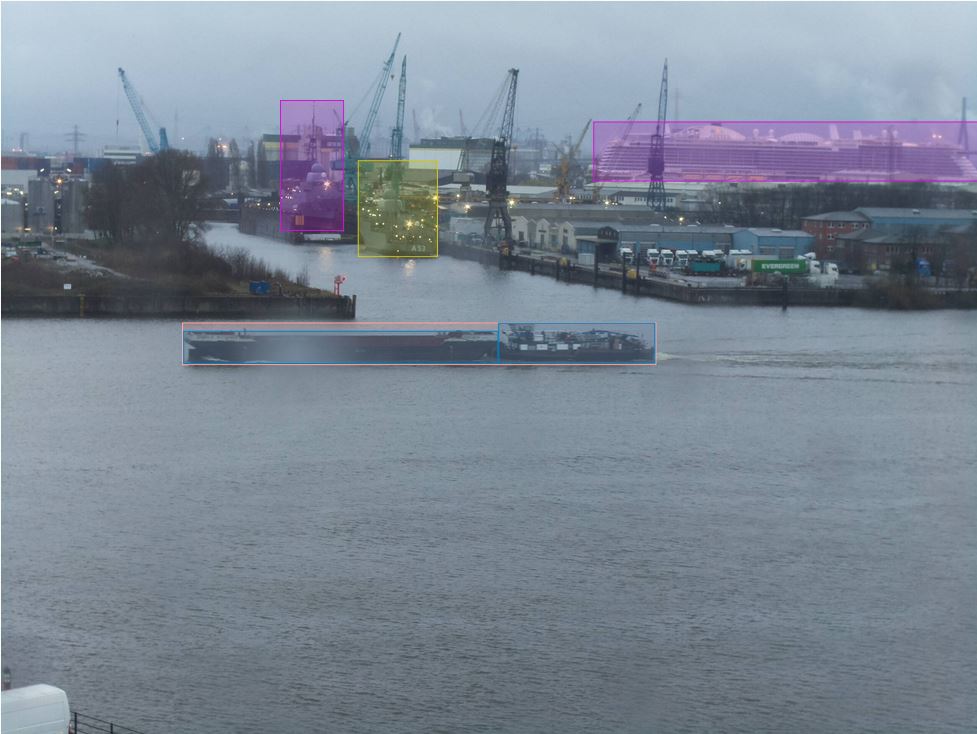}
    }
    \hfill
    \subfloat{
        \includegraphics[width=0.48\textwidth, trim=0cm 7cm 0cm 1cm, clip]{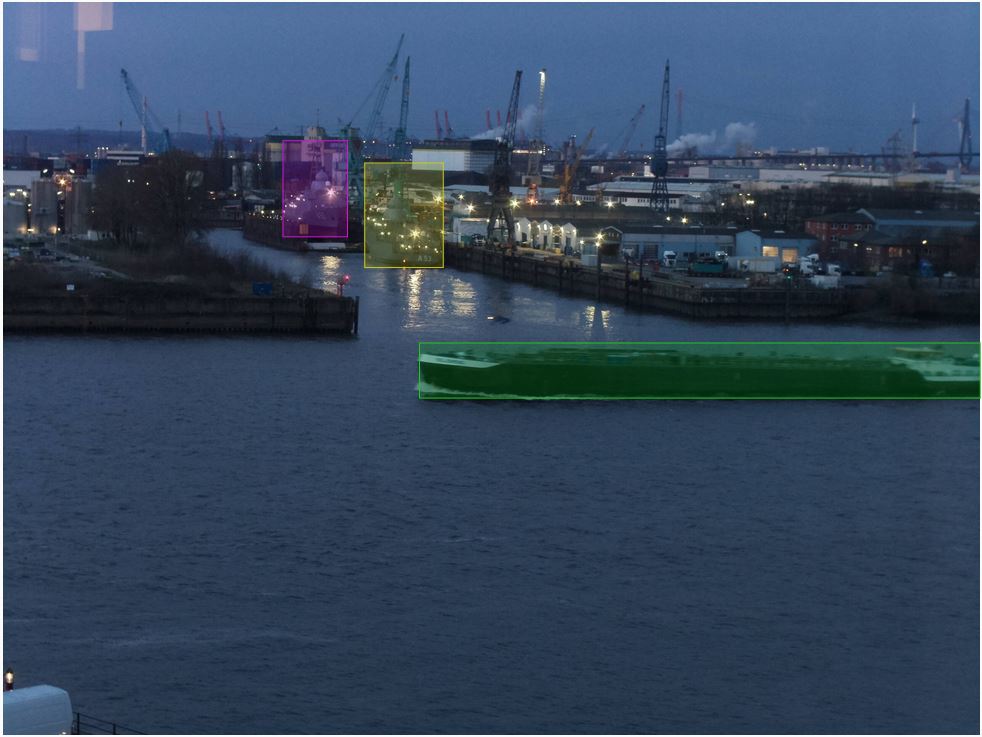}
    }
    \caption{Sample images from the dataset showing the 2D bounding box annotations with different sub-classes. Blue boxes highlight individual vessels in a barge-pusher setup, labelled as \texttt{subvessel}, while the entire setup is outlined in pink and labelled as \texttt{ship}. A cruise ship in the background or a ship in dry dock are marked with purple bounding boxes and labelled as \texttt{ship\_partial}. The yellow box indicates a docked ship, labelled as \texttt{ship\_moored}, and a green box shows a ship at the edge of the frame, labelled as \texttt{ship\_leaving\_frame}.}
    \label{fig:ObjDet_annotation_labels}
\end{figure*}

We also created a manually annotated set of 1000 images with object detection ground truth.
The object detection dataset has a intersection of images with the 6D pose estimation dataset, but is not a true subset.
To evaluate the object detection models on vessels, we annotated vessels with 5 sub-classes. 
 %: \texttt{ship}, \texttt{ship\_leaving\_frame}, \texttt{ship\_moored}, \texttt{ship\_partial} and \texttt{subvessel}. 
\cref{fig:ObjDet_annotation_labels} shows examples for all these classes. \texttt{ship} refers to a vessel cruising on the waterway, not touching the image edges. \texttt{ship\_leaving\_frame} is similar to \texttt{ship}, but it distinguishes vessels protruding beyond the image edge. \texttt{ship\_moored} is used for vessels that are fully visible, but moored against the quay wall in the background and not moving.  \texttt{ship\_partial} is reserved for ship-like objects, which are mostly occluded, in the background, and not moving. This includes vessels being built in the dock in the background, the superstructure of cruise ships mooring in the far background, and generally parts that are known to be of vessels, but very difficult to identify. The \texttt{subvessel} class is used for cases in which there is a configuration of multiple independent vessels travelling as a unit. For example, a pusher-barge unit is annotated as \texttt{ship} or \texttt{ship\_leaving\_frame} as whole and each part as a \texttt{subvessel}.

\section{Evaluation}
\label{sec:evaluation}
% In this section, we evaluate and discuss the performance of our technique across the key stages.

\subsection{Object detection results}
\label{sec:object_detection_results}
To select an object detection model for our system, we evaluated multiple detectors on our manually created object detection dataset of 1000 images. We evaluated different network sizes (X, L, S) of YOLOX alongside YOLOv3, as it is the basis for YOLOX. Furthermore we evaluate \ac{detr} \cite{carion2020end}, its evolution Deformable \ac{detr} \cite{zhu2020deformable}, and Cascade R-CNN with ResNet50 as backbone. % We also evaluated the other backbone variants of   Cascade R-CNN implemented in mmdetection, but found the performance increase negligible 

We will showcase results from two configurations, despite our annotation granularity allowing for more permutations. The first configuration involves using the classes \texttt{ship}, \texttt{ship\_leaving\_frame} and \texttt{subvessel} and the second configuration includes all of our sub-categories for vessels. Since the models we compare are pretrained on the MS COCO dataset, we compare these custom labels to the \texttt{boat} class.

We expect \ac{ais} messages for the three classes in the first configuration, as these vessels are moving, making them relevant for our system. This configuration allows us to evaluate networks especially for our use-case. Correspondingly, \texttt{ship\_moored} and \texttt{ship\_partial} are not relevant for our approach, as we expect no \ac{ais} data to be available for them. By including all classes in the second configuration, we provide a general overview for the ship detection performance of the object detectors. \cref{tab:DetectorEvalFastModelsMorePermutations} shows the performance of the detection models.

The results show that the combination of sub-classes \texttt{ship}, \texttt{ship\_leaving\_frame}, and \texttt{subvessel}, the relevant ones to our work, yields higher precision and recall then the one including all annotations. This suggests that \texttt{ship\_moored} and \texttt{ship\_partial} are harder to detect, as these vessels are often near quay walls and in front of a cluttered background. Especially \texttt{ship\_partial} includes many objects such as hulls not in the water or only the superstructures of cruise ships, features that differ significantly in appearance from a hull in water.

\begin{table*}
\centering
\resizebox{\linewidth}{!}{%
\begin{tabular}{l|l|*{8}{c}}
\toprule
\textbf{Ground truth} & \textbf{Metric} & \textbf{YOLOX-X} & \textbf{YOLOX-L} & \textbf{YOLOX-S} & \textbf{YOLOv3} & \textbf{DETR} & \textbf{Def. DETR} & \textbf{Def. DETR 2-Stage} & \textbf{Cascade R-CNN r50} \\
\midrule
\multirow{3}{*}{\makecell[l]{\textbf{Vessels relevant for our}\\ \textbf{use case (S, SLF, SV)}}}
& mAP@.5:.95 & \textbf{0.626} & 0.603 & 0.542 & 0.379 & 0.451 & 0.510 & 0.585 & 0.511 \\
& mAP@.5 & \textbf{0.805} & 0.794 & 0.764 & 0.659 & 0.696 & 0.737 & 0.775 & 0.725 \\
& AR@100 & 0.756 & 0.746 & 0.698 & 0.549 & 0.645 & 0.703 & \textbf{0.761} & 0.680 \\
\midrule
\multirow{3}{*}{\makecell[l]{\textbf{All vessel like objects}\\ \textbf{(S, SLF, SM, SP, SV)}}}
& mAP@.5:.95 & \textbf{0.398} & 0.363 & 0.277 & 0.232 & 0.278 & 0.337 & 0.372 & 0.309 \\
& mAP@.5 & \textbf{0.588} & 0.550 & 0.457 & 0.476 & 0.513 & 0.562 & 0.571 & 0.527 \\
& AR@100 & 0.514 & 0.473 & 0.429 & 0.332 & 0.398 & 0.447 & \textbf{0.531} & 0.427 \\
\bottomrule
\end{tabular}%
}
\caption{mAP and AR of various detectors on our ship detection dataset with 1000 images. The sub-class names are shortened for a better presentation. S: \texttt{ship}, SLF: \texttt{ship\_leaving\_frame}, SV: \texttt{subvessel}, SM: \texttt{ship\_moored}, SP: \texttt{ship\_partial}. YOLOX-X and the two-stage deformable DETR are the best performing models from the pool we evaluated. Since YOLOX-X was faster in inference by a factor of 2-3 on our hardware, we ended up choosing it over the DETR based one. The table also shows that the vessels relevant for our approach are easier to detect then the vessels in the background.}
\label{tab:DetectorEvalFastModelsMorePermutations}
\end{table*}

For comparing our results with existing research, we also used a model that was fine-tuned on a similar dataset\cite{gulsoylu2024image}. According to the results in \cite{gulsoylu2024image}, YOLOv5 achieved 0.332 mAP@0.5 without fine-tuning on their dataset. After fine-tuning on their dataset, the model's performance improved significantly to 0.951 mAP@0.5. However, on our dataset, the YOLOv5 model fine-tuned on their dataset performed similar to the generalised YOLOX-S, with mAP@0.5 of 0.455 considering the configuration with all sub-classes.
This shows that if we fine-tune a YOLOX-X model on our dataset, it may increase the performance on our dataset, but its transferability to other datasets will still be a question, especially as our scene is broadly static.

Therefore, to prove that our model can work with "off-the-shelf" models, we select the YOLOX-X model pretrained on MS COCO as the object detector we deploy to verify our approach. This model achieves 0.80 mAP@0.5 on relevant sub-classes, and sufficiently supports the proposed technique's performance. Moreover, the following section demonstrates that the object detection model does not significantly limit the performance.

% We can see that our dataset seems to be less complex than the one used in \cite{gulsoylu2024image}, as the older YOLOv3 performs significantly better on our dataset without fine-tuning, achieving 0.48 mAP@0.5 for all sub-classes.

\subsection{Projection error of georeferenced data into image space}
We evaluate the projection quality of the different projection methods via the average reprojection error of the annotated keypoints. We project all used world space keypoints into image space, then calculate the Euclidean distance between the transformed keypoints and the corresponding image keypoints. The camera was intermittently moved during data collection, leading to slightly changed viewports. For each transformation method, we calculate the average reprojection error of all keypoint correspondences over all viewports. 

\begin{table}[ht]
	\centering
    \resizebox{\linewidth}{!}{
	\begin{tabular}{l|ccc}
		\textbf{Transformation} & \textbf{MAE} & \textbf{$\frac{\text{Error}}{\text{width}}\%$} & \textbf{$\frac{\text{Error}}{\text{height}}\%$}  \\
		\hline
		\acf{pnp} & \textbf{12.22} & \textbf{0.48} & \textbf{0.64} \\
		Homography, Water surface & 61.88 & 2.42 & 3.22 \\
		Homography, \ac{pca} & 59.96 & 2.34 & 3.12 \\
		\hline            
		% Homography in \cite{gulsoylu2024image}, Altona & 39.97 &  3.12 & 5.51\\
		% Homography in \cite{gulsoylu2024image}, Blockbräu & 16.58 & 1.30 & 2.30\\
		% Homography in \cite{gulsoylu2024image}, Neumühlen & \textbf{11.39} & 0.59& 1.05\\
		Homography in \cite{gulsoylu2024image} & 22.65 & 1.67 & 2.95\\
	\end{tabular}
    }
	\caption{Resulting average projection errors for all transformation methods compared with \cite{gulsoylu2024image}. The results are displayed as the mean absolute error (MAE) in pixels, the absolute error relative to the image width, and the absolute error relative to the image height.}
	\label{tab:projection_quality}
\end{table}

In \cref{tab:projection_quality}, we compare the average projection error over our three projection methods. The \ac{pnp} method demonstrates a clear superiority, yielding a projection error that is one-fifth of the homography approaches. The differences between the two homography approaches are not significant. The \ac{pca}-based 2D projection plane improves the projection slightly compared to the water-surface-based projection plane. The higher error in the homography approaches can be explained by the loss of one dimension of information. Many of the keypoints differ in height, and as the camera views the surface plane at a relatively flat angle, these height differences have a significant impact on the reprojection error. The \ac{pca} approach performs marginally better as it is designed to minimise the information loss caused by discarding the height dimension. Nonetheless, the inability to represent the height information makes it significantly worse than the \ac{pnp} approach. 

\cref{tab:projection_quality} also compares our results to those from \cite{gulsoylu2024image}, where they used homography for registering the image in the real world. Our \ac{pnp} approach outperforms the homography-based approach in \cite{gulsoylu2024image}. The performance of homography error relative to image size is slightly worse in our case. This discrepancy could be due to variations in the distribution of keypoints along the world axes. Selecting 3D keypoints that are near a plane, results in minimal information loss for homography. Conversely, if the keypoints are spread out across all three dimensions, the reprojection error tends to be larger.

Choosing real-world keypoints only on a plane enhances the average projection error metric, yet it does not address the inherent limitation of homography to project points not situated on the keypoint plane. Therefore, we demonstrate the superiority of employing the \ac{pnp} method for converting between image and world space. Our approach requires \ac{pnp} for the complete camera model and create 3D bounding boxes. However, related work on matching could also gain from the comprehensive three-dimensional mapping achievable with a perspective projection model, thereby improving the precision of projection when points are not coplanar.

\subsection{Analysis of \ac{ais} and detection matching}

\begin{table*}[]
	\centering
	\begin{tabular}{l l r r r}
		\toprule
		\textbf{Association result} & \textbf{Quality/failure reason} & \textbf{Count} & \textbf{(\%) over Grand Total} & \textbf{(\%) in final dataset} \\
		\midrule
		\multirow{4}{*}{Correct match}
		& Good 3D bounding box & 375 & 64.54 & 86.41 \\
		& OK 3D bounding box & 33 & 5.68 & 7.60 \\
		& Bad 3D bounding box & 2 & 0.34 & 0.46 \\
		& \textbf{Total} & \textbf{410} & \textbf{70.56} & \textbf{94.47} \\
		\midrule
		\multirow{5}{*}{Wrong match}
		& Due to missing \ac{ais} data & 9 & 1.55 & 2.07 \\
		& Due to false positive detection & 2 & 0.34 & 0.46 \\
		& Due to cost function & 9 & 1.55 & 2.07 \\
		& Subvessel and combined unit confusion & 4 & 0.69 & 0.92 \\
		& \textbf{Total} & \textbf{24} & \textbf{4.13} & \textbf{5.52} \\
		\midrule
		\multirow{5}{*}{No match}
		& Vessel not predicted & 3 & 0.52 & \\
		& Vessel has no \ac{ais} & 139 & 23.92 & \\
		& Vessel has neither prediction nor \ac{ais} & 2 & 0.34 & \\
		& Has match and \ac{ais}, but cost was too high & 3 & 0.52 & \\
		& \textbf{Total} & \textbf{147} & \textbf{25.30} & \\
		\bottomrule
	\end{tabular}
	\caption{Quality of Matching and 3D Bounding box creation of our system for  581 vessels across a sample of 500 images, categorised by match result (Correct match, Wrong match, No match). Subcategories show 3D bounding box quality or matching failure reasons. The final columns show the case distributions from two different perspectives: From the initial image set, over all vessels relevant for our system and from the final 6D pose estimation dataset resulting from our approach, where "No match" cases are excluded as they yield no annotations. This can be read as e.g.: From 100 vessel occurrences fed into our system, 71 were successfully associated with \ac{ais} messages; From 100  annotated vessels in our dataset, 94 annotations are based on a correct match.}
	\label{tab:vessel_matching_results}
\end{table*}

% \textit{This table is the result of the first manual eval run I did with 500 images of the final 6D pose dataset. The IoU distribution was created with a newly sampled set: All 1000 images for which 2D BB GT exists where fed through the system, and all of those for which poses could be created where evaluated with categories. Those where ~250 images (2D BB GT was sampled randomly during its creation, many of those where in the timeframe for which no AIS data does exist. This now leads to this table and the IoU distribution evaluation later having completely different results. In the manual eval run for the 250 images in the IoU Dist i was way stricter with the evaluation.}

To evaluate the performance of the matching algorithm, we analyse a randomly selected subset of 500 images from our dataset. In these 500 images, a total of 581 vessels are visible. \cref{tab:vessel_matching_results} presents the distribution of results for matching vessel occurrences in the images with the \ac{ais} data, categorising them into correctly matched, incorrectly matched, or unmatched. For each image, we verified whether each vessel was successfully detected. We then checked the availability of \ac{ais} data for each detected vessel and manually assessed whether our matching algorithm correctly associated the 2D bounding boxes with their corresponding \ac{ais} data. Our matching algorithm achieves an overall accuracy of 70.56\%, which is comparable to related work such as Lu et al.'s framework \cite{lu2021fusion}, obtaining 75.70\% accuracy with \ac{dr}, Huang et al.'s framework \cite{huang2021identity} achieving 74.3\% accuracy on average, Qu et al.'s framework \cite{qu2022intelligent} with an overall accuracy of 81.42\% and Gülsoylu et al.'s framework \cite{gulsoylu2024image} with an average accuracy of 74.76\%.

A correct match occurs when there is a predicted 2D bounding box around the vessel and it is correctly associated with the \ac{ais} data of the vessel in the bounding box. This is a prerequisite for creating a correct 3D bounding box for the vessel in the data fusion stage of our system. "Wrong match" occurs when the matching algorithm associates a vessel in a 2D bounding box with \ac{ais} data  from a different vessel. This results in a 6D annotation that is based on incorrect \ac{ais} data, meaning that the vessel is annotated with e.g. wrong size or heading. A wrong match has the most detrimental effect on the quality of our dataset.

Wrong matches can occur when a false positive 2D bounding box gets matched with the \ac{ais} data of a nearby vessel, leading to a incorrect 3D bounding box creation for the actual vessel. Confusion between subvessels in multi-vessel units is another issue. If the \ac{ais} data describes only a part of the unit that is not visible, mismatched associations can occur, resulting in incorrect 6D pose annotations. Wrong matches may also stem from missing \ac{ais} data for the correct vessel. In such cases, the system matches detections with available \ac{ais} data from other vessels, leading to errors. Lastly, limitations in the cost function can lead to incorrect matches. Even when both detection and \ac{ais} data for the correct vessel are present, the correct match can be forgone when the cost function selects another match with a lower cost.

"No match" refers to a vessel that is visible, but not detected or not associated with any \ac{ais} data by the bipartite graph matching.
The most common reason for no match is the lack of \ac{ais} data. This can be caused by the vessel not transmitting \ac{ais} or the messages being missed by our capturing. Missing heading information of a vessel also makes it impossible to create an oriented water-level plane segment for the vessel, leading to the "No AIS" case. When the object detector fails to detect a visible vessel, this prevents a match from being made despite accurate \ac{ais} data. In some instances, neither a detection nor \ac{ais} messages exist for the vessel. 
% There are also cases where the design of the cost function prevents correct matches with both predictions and \ac{ais} present from being made.

Overall, the availability of \ac{ais} data is the biggest limitation for correct matching, which can either prevent matching altogether or lead to inaccurate matches. In both cases, the correct match can not be created due to missing data, which then leads to no match being created or a wrong match being chosen instead. There is still improvement possible for the cost function and a better one could improve the fraction of correct matches in \cref{tab:vessel_matching_results} further. 
 
\subsection{Analysis of AIS data correction quality}
\begin{figure*}[t]
    \centering
    \subfloat{\includegraphics[width=0.32\textwidth]{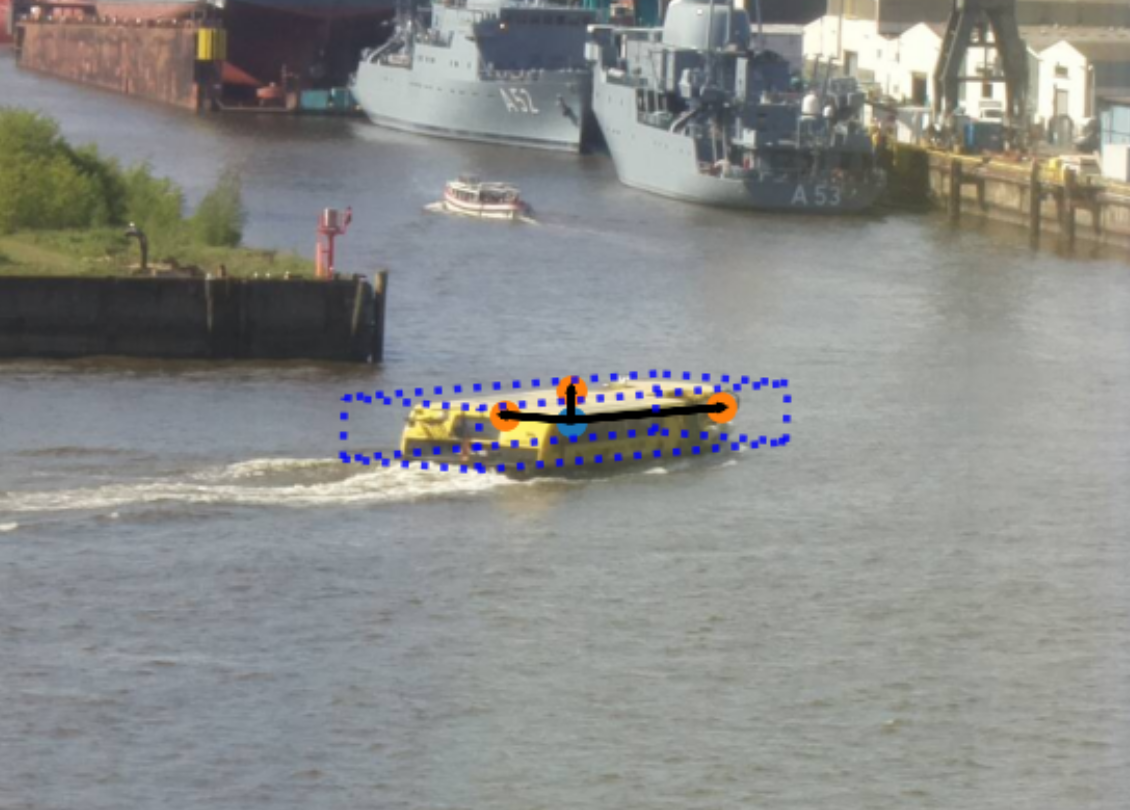}}
    \hfill
    \subfloat{\includegraphics[width=0.32\textwidth]{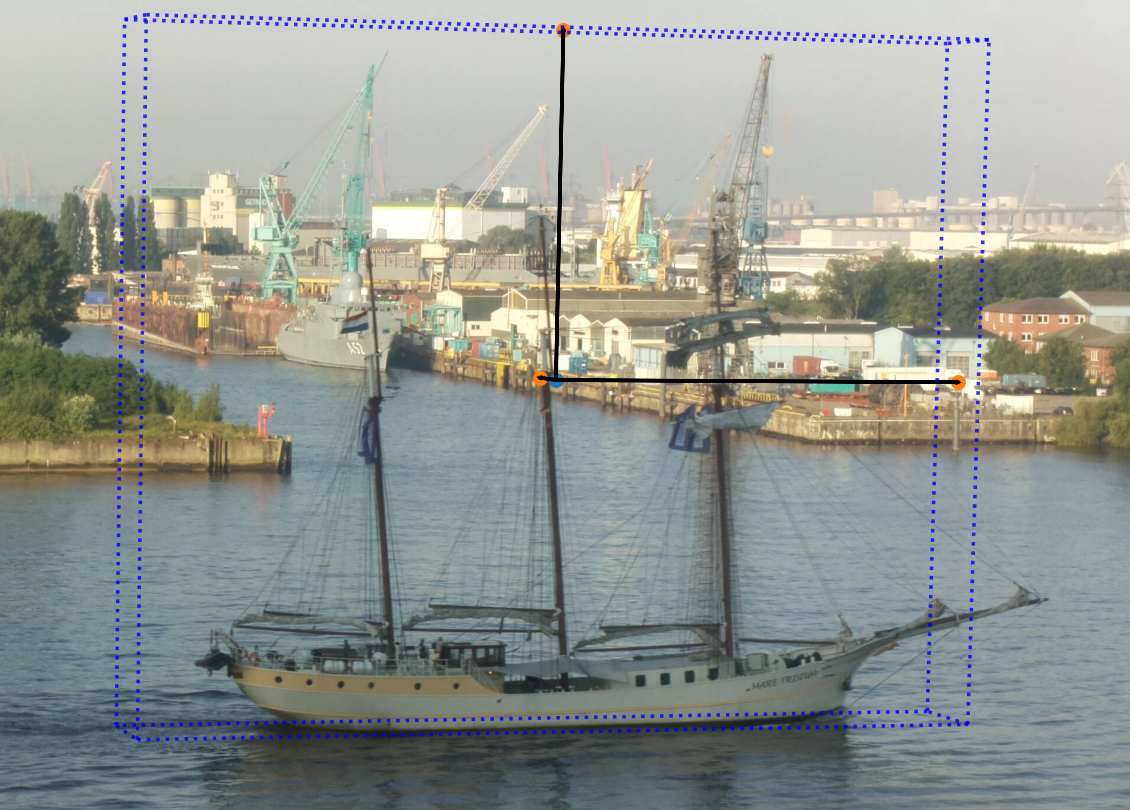}}
    \hfill
    \subfloat{\includegraphics[width=0.32\textwidth]{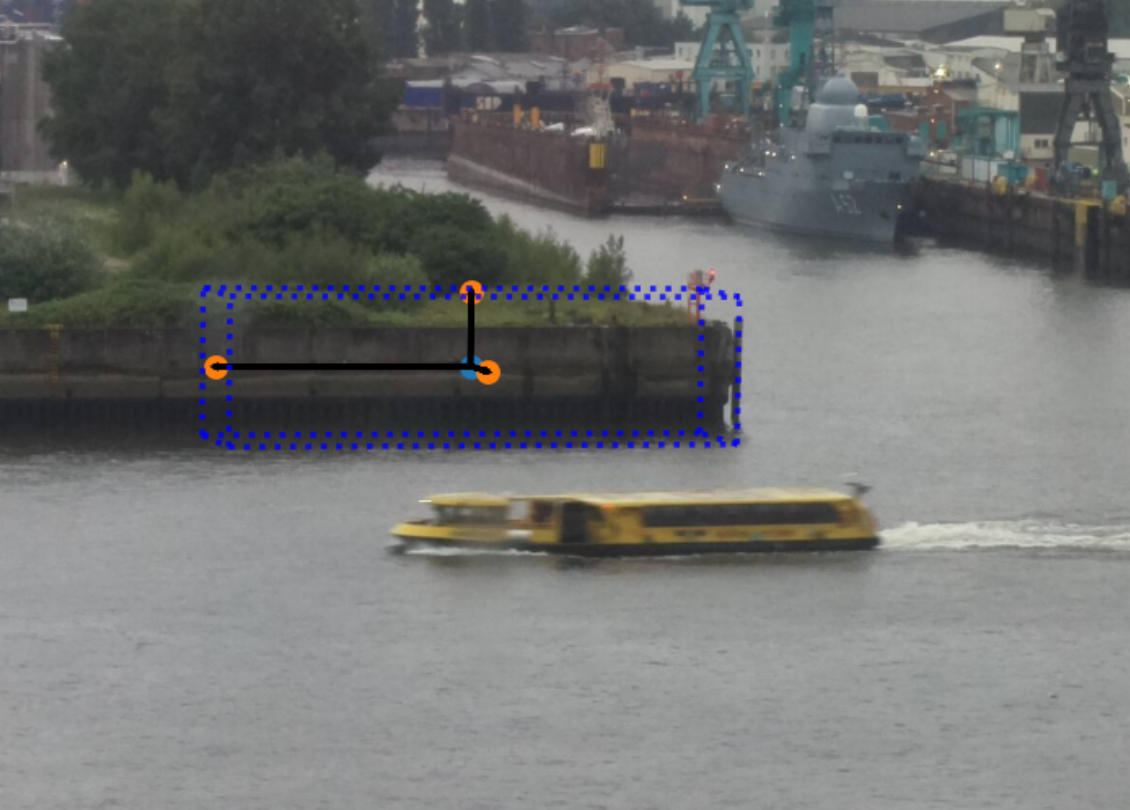}}
    \caption{Examples of non "Good" annotations produced by our system. The left most one is an "OK" annotation, there is a mismatch between the heading in the \ac{ais} and in the actual rotation of the vessel visible. 
    % In our \ac{iou} proxy metric, this would still achieve a high score.
    The middle one shows the object detector getting confused by the sailing boats masts and boat-spirit. This causes the 3D bounding box to be both offset to the left and much higher than it should be. It is a "Bad" bounding box.
    The right one shows a \ac{ais} based water-level plane segment being associated with a false positive detection, a "Wrong match" failure. This causes the 6D pose to be annotated in a region not containing a vessel.}
    \label{fig:PoseQualityExamples}
\end{figure*}

For the 581 vessels visible in the 500 images we sampled, our technique produced 410 matches, each leading to a 3D bounding box to be created. For each of those associations, we manually evaluated the quality of the resulting bounding box and categorised them into categories "Good", "OK", "Bad" and "Wrong match", as shown in \cref{tab:vessel_matching_results}. This evaluation relied on manual effort.

"Good" annotations are those where the 3D bounding boxes precisely align with the actual pose of the visible vessel. These annotations accurately represent the vessel's position and orientation even in complex scenarios involving multiple vessels or partial occlusions. Examples for these are shown in \cref{fig:PoseDatasetSamples}.
"OK" annotations are 3D bounding boxes that mostly fit the vessel but may have slight inaccuracies, such as an off-centre location or an skewed heading. An example is a vessel manoeuvring in front of the camera. If the vessel is mid-turn, it may have a different heading in the image compared to the one transmitted in an outdated \ac{ais} message. This discrepancy causes the bounding box to be slightly rotated against the actual heading. This is shown in the left image in \cref{fig:PoseQualityExamples}.
"Bad" annotations are characterised by a significant deviation between the created 3D bounding box and the visible vessel, despite a correct association. For example, a sailing ship might be visible, but the object detector outputs a bounding box that includes background clutter as its masts. This causes the height inference to produce a bounding box that is much to big, an example shown in the middle of \cref{fig:PoseQualityExamples}.
For a "Wrong match", the 3D bounding box is always incorrect, often not fitting the dimensions, location, and heading of the visible vessel. A  failure, resulting in a completely erroneous annotation, is show in the right image in \cref{fig:PoseQualityExamples}.

\begin{figure}[]
	\centering
	\includegraphics[width=\linewidth, ]{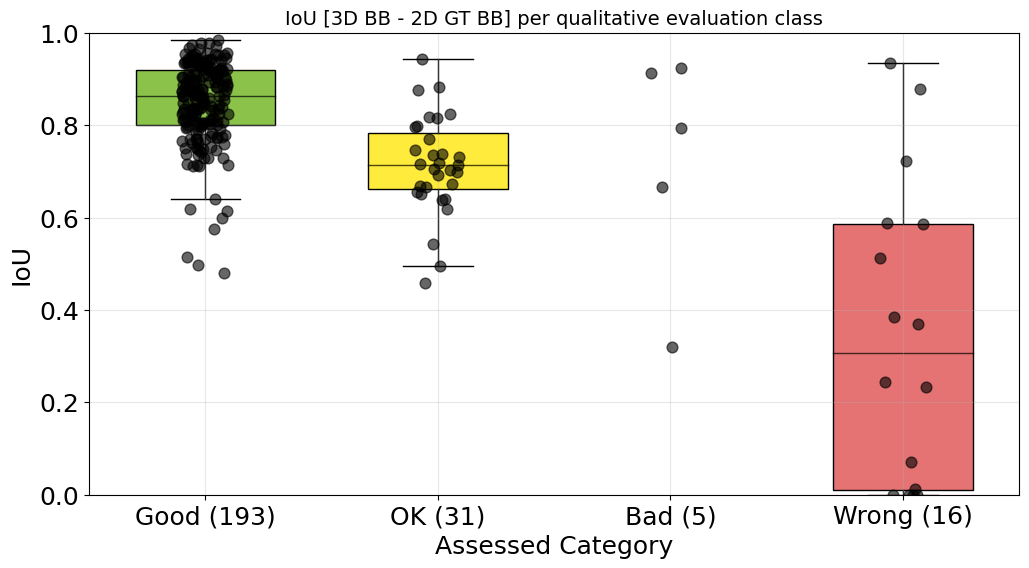}
	\caption{\ac{iou} distribution of 2D bounding boxes and created 3D bounding boxes, separated by categories of manual quality evaluation.}
	\label{fig:ObjDet_annQual_IoU}
\end{figure}

Since there is no ground truth for the height information, we cannot evaluate the output of our technique with commonly used 6D pose estimation metrics such as \ac{add-s} \cite{hinterstoisser2012model}. However, to put these quality categories into a quantitive frame of reference, we calculated the \ac{iou} of a vessel's manually annotated 2D object detection bounding box and the minimal enclosing rectangle of the 3D bounding box created by our system. This \ac{iou} compares the rectangular 2D silhouette of our automatically created 3D bounding box in image space with the human annotated rectangular 2D vessel object ground truth. This measure is sensitive to spatial inaccuracies but less sensitive to rotational errors. \cref{fig:ObjDet_annQual_IoU} shows the distribution of our \ac{iou} measure categorised by the human assessment of the annotation quality. This calculation was performed on 245 images that are present in both the set of 1000 images annotated for object detection and in the final dataset of 3753 images produced by our approach. 
% This intersection is unfortunately rather small, as there is no captured \ac{ais} available at all for a large part of the images in our 2D object detection dataset, leading to them not even being included in the 4176 images fed into our system.

We see that our quality assessment of the 3D bounding boxes, created by our system, correlates with the overlap between their projections in image space and the 2D object detection bounding boxes. We manually analysed the outliers that did not follow our expectation of "Good" annotations having a high \ac{iou} and "OK" and "Bad" ones having a lower \ac{iou}. We found that the samples with high \ac{iou} scores in the "OK" or "Bad" categories are not classified as "Good" due to heading errors. A 10$^{\circ}$ heading error has a low impact on the projection of the 3D bounding box and does not impede a high \ac{iou} score. In our quality assessment such a large heading error prevents classification as "Good". The lower \ac{iou} outliers in the "Good" category are due to differences in how antennas on vessels are handled. For example, if a 3D bounding box does not include the superstructure (e.g. antennas) fully due to mistakes in 2D bounding box prediction, the \ac{iou} is lower without necessarily making the pose not "Good". 

% We also looked of the distribution of the correction vector $\Delta \mathbf{v}_{3d}$. When transformed into image space, we see that the distribution has a clear peak around \((0, 0)\), showing that in most cases, the created correction has a low magnitude. The average vertical magnitude is greater then the horizontal one. As most ships cross the viewport travelling from one side to the other, the sideway movement could be explained by the prediction correcting a \ac{ais} position guess in the direction of travel, correcting for a inaccurate \ac{dr} position estimation. The vertical correction could correct for changes in tide, as tide is not part of the \ac{ais} based location prediction but influences the vessels actual position in the image.

Even though finding precise metrics to evaluate the performance of our system regarding the quality of the final 6D pose annotations is hard, without having objective 6D ground truth annotations already available, we have shown that there is a high percentage of good 6D pose estimation annotations created by our approach.

Missing \ac{ais} data is the most impactful factor reducing the share of correct and accurate annotations. 
% Using a more complete database of \ac{ais} messages, then what was available to us, can improve the performance of the system we introduced significantly.
Wrong matches are a significant source of incorrect 6D poses in the final dataset. Improvements in the cost function of the bipartite graph matching could increase the share of accurate annotations further, even as the current parametrisation is a local optimum devised in our experiments. 
%@Emre added end of sentance to avoid "Why didn't you tweak function further?" question
The "OK" bounding boxes are caused by either a wrong 2D bounding box prediction or inaccurate heading information from the \ac{ais} data. The former could be improved by fine-tuning the detector, the latter by a more sophisticated interpolation of the heading, by taking its rate of turn, and course into account.
 
\section{Conclusion}
%Recap
In this paper, we presented a novel technique to associate vessel detections on monocular 2D RGB images with \ac{ais} data for generating accurate 3D bounding boxes. Methods for coordinate transformation between \ac{gps} coordinates and image coordinates were compared, showing that a \ac{pnp}-based approach achieves lower projection errors than previously used homography based ones. Multiple object detectors were benchmarked on a subset of our collected images, with YOLOX-X selected for its favourable speed-accuracy balance.

The matching problem was modelled as bipartite graph matching and the evaluation showed that the matching algorithm achieves an accuracy of 70.56\%, which is comparable to related work \cite{lu2021fusion, qu2022intelligent, gulsoylu2024image}. The most common reason for incorrect matching is the lack of \ac{ais} data. Collecting Class B \ac{ais} data along with Class A could lead to better association.

%Should this be included? Somewhere else maybe in the eval section?
%Looking at using \ac{ais} matching to reduce false positives in vessel detection, the matching successfully reduces false positives with only two accepted false positive detections in a test of 500 images. But it also excludes about 30\% of true positives due to missing \ac{ais} signals, making the approach highly selective and completely dependand on the presence of \ac{ais}-data.

Additionally, the paper introduces the \ac{bonk-pose}, a publicly available dataset comprising 3753 annotated images, which can be used for training and evaluating ship detection and 6D pose estimation networks. The quality of the final dataset is assessed manually and with the IoU of 2D bounding boxes and 3D bounding boxes in the image space. For vessels with successful matches, 6D poses were obtained by refining the \ac{ais}-based pose using visual cues derived from the monocular RGB images. The 6D pose annotation dataset created by our approach, without any manual posterior filtering, contains 86.4\% correct annotations and 94.5\% annotations of at least acceptable quality (see \cref{tab:vessel_matching_results}).
%@emre remove repetition of future work
%, and we are confident that this percentage can be further increased by optimising the cost function and the availability of \ac{ais} data.
%\todo{Revise percentages after eval is cleaned up}

%Future work
Overall, the proposed technique demonstrates that fusing \ac{ais} data with object detections is a viable approach for creating accurate 6D pose annotations of vessels in maritime imagery. To further enhance the technique's performance, future work could focus on several improvements. Fine-tuning the object detection model can improve the detection quality. 
% Although not many incorrect or missing matches are caused by object detection model, inconsistent handling of superstructures in bounding box height causes 6D poses to have incorrect height estimations.

% While our approach created some great results on our collected images, the nature of how they were collected causes a static scene, greatly reducing their diversity. Especially the uneven distribution of heading in vessels passing by the camera causes the rotations of our vessel poses to be highly skewed.

A way to automate the registration of the viewport in the real world would be beneficial, to avoid the manual keypoint annotation that was necessary to solve the camera pose in the real world. Adapting this technique for a mobile setup with a camera and \ac{ais} antenna, would pave the way for a diverse image collection. 

Finally, porting our approach to work on temporal image series could improve the matching accuracy, by combining an object tracker. Tracking the vessels would provide more information such as the movement direction and speed. These information can be auxiliary to the matching cost function. 
% Kiefer et al. \cite{kiefer2023stable} showed that for heading estimation, time series can lead to more stable estimation results, Huang et al. \cite{huang2021identity} used image tracks to improve the matching accuracy to \ac{ais}.

\bibliographystyle{IEEEtran.bst}
\bibliography{references}

\end{document}